%% file: bare_jrnl.tex
\documentclass[journal]{IEEEtran}

\usepackage[utf8]{inputenc} 
\usepackage[T1]{fontenc}    
\usepackage[pagebackref=true,breaklinks=true,letterpaper=true,colorlinks,bookmarks=false,citecolor=blue,linkcolor=red]{hyperref}       
\usepackage{url}            
\usepackage{booktabs}       
\usepackage{amsfonts}       
\usepackage{nicefrac}       
\usepackage{microtype}      
\usepackage{bm}
\usepackage{graphicx}
\usepackage{subfig}
\usepackage{sidecap}
\usepackage{algorithm, algpseudocode}
\usepackage{amsmath}
\usepackage[bb=boondox]{mathalfa}
\usepackage{url}

\usepackage{caption}
\usepackage{enumitem}
\usepackage{multirow}
\usepackage{color}
\usepackage{xspace}

\usepackage{amssymb}
\usepackage{pifont}

\usepackage{color}
\usepackage{colortbl}
\usepackage{textcomp}
\usepackage{stmaryrd}
\usepackage{wrapfig}
\usepackage{caption}
\usepackage{enumitem}
\usepackage{pifont}  
\usepackage{array}
\usepackage{enumitem}
\usepackage{pifont}
\usepackage{sidecap}

\newcommand{\PreserveBackslash}[1]{\let\temp=\\#1\let\\=\temp}
\newcolumntype{C}[1]{>{\PreserveBackslash\centering}p{#1}}
\newcolumntype{R}[1]{>{\PreserveBackslash\raggedleft}p{#1}}
\newcolumntype{L}[1]{>{\PreserveBackslash\raggedright}p{#1}}

\definecolor{Gray}{gray}{0.90}
\definecolor{LightCyan}{rgb}{0.82,0.82,1}

\newcolumntype{a}{>{\columncolor{Gray}}c}
\newcolumntype{b}{>{\columncolor{LightCyan}}c}

\usepackage{array}

\usepackage{lineno,hyperref}
\usepackage{xspace}
\newcommand{\etal}{\emph{et al.}\xspace}
\newcommand{\eg}{\emph{e.g.}\xspace}
\newcommand{\ie}{\emph{i.e.}\xspace}
\modulolinenumbers[5]

\begin{document}
\title{Guidance Through Surrogate: Towards a Generic Diagnostic Attack}

\author{Muzammal Naseer,~Salman Khan,~Fatih Porikli and~Fahad Shahbaz Khan
\thanks{M. Naseer is with the Mohamed Bin Zayed University of Artificial Intelligence (MBZUAI), and Australian National University (ANU), Canberra AU, e-mail: (\texttt{\footnotesize muz.pak@gmail.com}).}
\thanks{S. Khan is with Mohamed Bin Zayed University of Artificial Intelligence (MBZUAI), UAE and Australian National University (ANU), Canberra AU.}
\thanks{F. Porikli is with Qualcomm, USA.}
\thanks{F. S. Khan is with the Mohamed Bin Zayed University of Artificial Intelligence (MBZUAI), and Link\"{o}ping University, Sweden.}
}

\markboth{Journal of \LaTeX\ Class Files,~Vol.~14, No.~8, August~2015}%
{Shell \MakeLowercase{\textit{et al.}}: Bare Demo of IEEEtran.cls for IEEE Journals}

\maketitle

\begin{abstract}
Adversarial training is an effective approach to make deep neural networks robust against adversarial attacks. Recently, different adversarial training defenses are proposed that not only maintain a high clean accuracy but also show significant robustness against popular and well studied adversarial attacks such as PGD. High adversarial robustness can also arise if an attack fails to find adversarial gradient directions, a phenomenon known as `gradient masking'. In this work, we analyse the effect of label smoothing on adversarial training as one of the potential causes of gradient masking. We then develop a guided mechanism to avoid local minima during attack optimization, leading to a novel attack dubbed \emph{Guided Projected Gradient Attack} (G-PGA). Our attack approach is based on a \emph{`match and deceive'} loss that finds optimal adversarial directions through guidance from a surrogate model. Our modified attack does not require random restarts, large number of attack iterations or search for an optimal step-size. Furthermore, our proposed G-PGA is generic, thus  it can be combined with an ensemble attack strategy as we demonstrate for the case of Auto-Attack, leading to efficiency and convergence speed improvements. More than an effective attack, G-PGA can be used as a diagnostic tool to reveal elusive robustness due to gradient masking in adversarial defenses. 
\end{abstract}


\IEEEpeerreviewmaketitle

\section{Introduction}
\IEEEPARstart{A} defense can cause gradient masking if it does not allow an adversarial attack to calculate useful gradient directions to deceive a model. Papernot \etal \cite{papernot2017practical} found that gradient masking alone is not a robust way to devise a well-rounded defense since adversarial perturbations can be discovered for such models using alternative means \eg a smooth version of the same model or from a substitute model. In this manner, the same attack can be used to fool the model by intelligently estimating perturbation directions to which the model remains highly sensitive even after deploying the defense. 

We hint towards the presence of gradient masking in the recent state-of-the-art defense mechanisms (Feature Scattering \cite{feature_scatter},  AvMixup \cite{lee2020adversarial} and Mixup-Inference \cite{Pang2020Mixup}). These defenses show excellent robustness in the more challenging white-box setting, where all network parameters, model architecture and training details are known. Specifically, single-step~\cite{goodfellow2014explaining,tramer2017ensemble} as well as iterative optimization-based \cite{madry2018towards,Kurakin2016AdversarialEI,xie2019improving,dong2018boosting} attacks find it difficult to calculate useful gradient directions in order to launch a successful attack. We note that these defenses use \emph{label smoothing} \cite{muller2019does,pereyra2017regularizing} as a regularization measure to create smooth loss surfaces. While learning smooth loss surfaces is a preferable property of robust models \cite{Zhang2019theoretically}, the existence of large contagious regions of adversarial examples \cite{tabacof2016exploring,goodfellow2014explaining} means the smoothness is only achieved in the small neighborhood of the training manifold. We demonstrate that label smoothing can thus cause gradient masking, thereby leading to inflated estimates of model robustness.

\begin{figure}[!t]
    \centering
    \includegraphics[width=0.9\columnwidth]{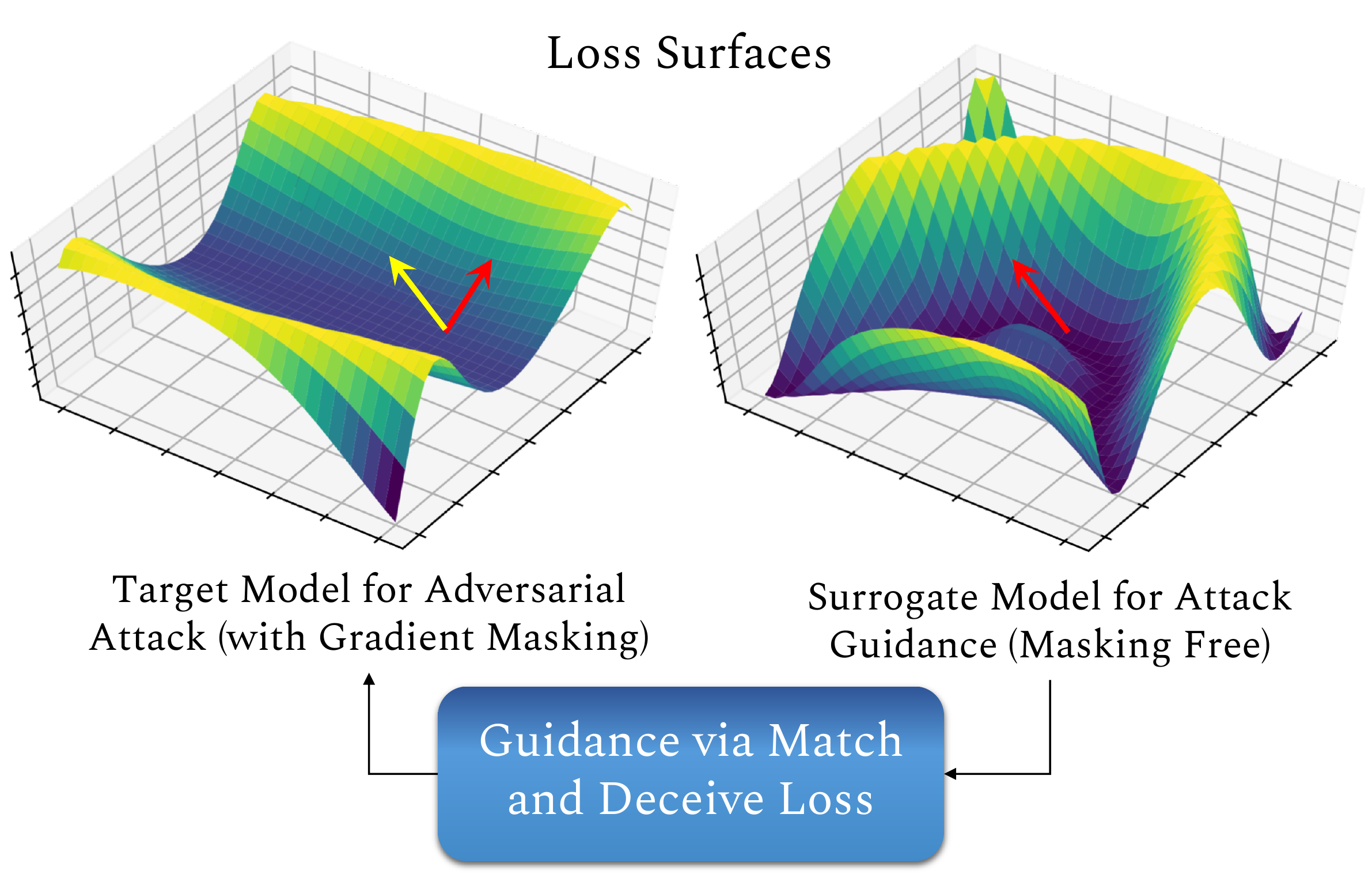}
    \caption{\centering We analyze the effect of gradient masking on popular defenses and propose a guidance based attack algorithm, called G-PGA. Our approach seeks guidance from a masking-free surrogate model to find useful attack directions which in turn can be used to diagnose gradient masking.}
    \label{fig:overview}
\end{figure}

In this work, by studying the masking behaviour of label smoothing, we characterize a new category of gradient masking which may not be intentionally caused with shattered, randomized or vanishing gradients. Instead, we suggest the specific `\emph{loss surface}' induced by a loss function and/or the  training algorithm can also lead to gradient masking. This means a white-box attack would not be much successful even with all the knowledge about training process and network parameters.  We found that since such a masking is quite subtle, it may not be caught by the diagnostic behaviours identified by Athalye \etal \cite{athalye2018obfuscated}. Subsequently, we propose a new attack approach (G-PGA) based on the adversarial directions from a surrogate model that acts as a useful test for existing defenses. Our approach utilizes a novel contrastive `\emph{match and deceive}' loss to find harmful directions using guidance from the teacher to deceive the source model (Fig.~\ref{fig:overview}).

\textbf{Contributions:} The main contribution of this work is to study label-smoothing as  a case-study to highlight the masking behaviour of recent popular defenses. We complement our findings with a strong attack method that generalizes across various defense mechanisms to unveil the gradient masking effect. The salient aspects of our approach are,
\begin{itemize}
    \item \emph{\textbf{Relation of Label Smoothing with Gradient Masking:}} We analyse the effect of label smoothing on adversarial training (AT). To this end, we develop an AT algorithm, called mask-AT, that combines random initialization and label smoothing with a single step adversarial attack to achieve masking without losing clean accuracy. Our mask-AT highlights label smoothing as the common root cause of elusive robustness of the recent defenses \cite{feature_scatter, lee2020adversarial}.
    \item \emph{\textbf{Guiding Mechanism:}} We develop a guiding mechanism that allows finding useful gradient directions during attack optimization. The idea is to exploit gradient information from a \emph{masking-free} surrogate model to distill and strengthen the adversarial noise. This way attack optimization avoids being stuck to local minima by observing correct gradient direction.  The masking-free surrogate provides healthy gradient directions \cite{athalye2018obfuscated} e.g., naturally or adversarially trained models  (Madry \cite{madry2018towards} or Trades \cite{Zhang2019theoretically}).
    \item \emph{\textbf{Match and Deceive loss:}} Our guided attack called G-PGA is based on a novel `\emph{match and deceive}' loss function. The purpose of this loss is to exploit surrogate information based on the principals of rescaling and redirection, achieved by normalized cross-entropy and contrastive directional objectives, respectively  (Sec.~\ref{sec:methodology}).
\end{itemize}

\section{Related Work}
\textbf{Adversarial Training:} Robust training \cite{madry2018towards, Zhang2019theoretically, shafahi2019adversarial, zhang2019you, zhang2019towards, zhang2019towards, salman2019provably} constitutes of finding adversarial examples by maximizing the model loss and then  updating model parameters to correctly classify them. Many robust training methods are proposed including Madry \etal\cite{madry2018towards} that solved the maximization step with an iterative and computationally expensive attack. Recently, \cite{Wong2020Fast} reduced the training cost significantly with a single step attack \cite{goodfellow2014explaining} combined with better initialization and early termination. Zhang \etal \cite{Zhang2019theoretically} proposed to control the trade-off between clean and adversarial accuracy. Adversarial robustness was further improved in  \cite{wang2020improving} by focusing on the misclassified examples during training. Carmon \etal\cite{carmon2019unlabeled} proposed to train on extra unlabeled data that lead to higher robustness with lower drop in clean accuracy. 
Zhang \etal \cite{feature_scatter} find adversaries by maximizing optimal transport distance and achieved high robustness when combined with label smoothing. Similarly, \cite{lee2020adversarial} further enhances the adversarial robustness with minimal loss in clean accuracy. They interpolate between clean and adversarial samples as well as labels with different smoothing factors.  Another line of defenses use input processing  \cite{guo2018countering, Akhtar_2018_CVPR, naseer2020self} to mitigate the adversarial effect. For example \cite{Pang2020Mixup} exploits mixup \cite{zhang2017mixup} based processing along with adversarial training to achieve higher robustness.

\textbf{Gradient masking:} Previous work from Athalye \etal \cite{athalye2018obfuscated} noted three causes for obfuscated gradients namely shattered gradients, stochastic gradients and vanishing/exploding gradients. For the first problem they propose BPDA, EOT attack for the stochastic/randomized defense and reparametrization and stable optimization for the third category. Further, they reported a number of tests to characterize gradient masking such as one step attacks must perform inferior to iterative attacks and black-box attacks should perform lower than white-box attacks. However, the identified behaviors do not form a complete set of possible indicators for gradient masking. In this paper, we study state of the art defense approaches \cite{feature_scatter,lee2020adversarial} and note that although they generally pass the tests prescribed by \cite{athalye2018obfuscated}, they are still potentially suffering from gradient masking. This is demonstrated by our extensive experiments that show a significant drop in their performance under our attack.

\textbf{Adversarial Attacks:} A number of adversarial attacks \cite{papernot2016limitations, modas2019sparsefool, su2019one,  carlini2017towards, naseer2019cross, Naseer_2021_ICCV, naseer2019local, chen2018ead, wu2020skip, naseer2021improving}  have been proposed to evaluate the robustness of deep neural networks. Among them, Projected Gradient Descent (PGD) \cite{madry2018towards} and CW \cite{carlini2017towards} attacks are computationally feasible and the most popular ones. However, these attacks fail to faithfully estimate robustness of recent defenses \cite{feature_scatter, lee2020adversarial} even when they are adaptive to the defense \cite{Pang2020Mixup}. This has lead to the development of new attacks where \cite{tashiro2020output} introduced a sampling strategy to enhance performance of PGD and CW, while  \cite{croce2020reliable} proposed Auto-Attack based on parameter free objectives along with evaluating the model on ensemble of attack strategies. Auto-Attack depends on the large number of queries, and extra information such as identifying which input samples are missclassified by a model and then adapt to the new attack settings for those input samples. In real-world settings, this information may not be available to the attacker.

\textbf{Our Differences:} Our proposed guiding mechanism is generic in its nature. When used as a stand-alone attack, it performs equivalent to state-of-the-art Auto-Attack \cite{croce2020reliable}  while being significantly less computationally expensive. It does not require large number of queries or random-restarts. It also does not depend upon extra information such as miss-classification indication for a given sample. Our proposed method complements the current attacks. When our guided mechanism is combined with existing approaches \cite{croce2020reliable, andriushchenko2019square, croce2020minimally}, it allows faster convergence and enhances the performance of an attack as shown in combination with attack strategies proposed by Auto-Attack in Sec. \ref{sec:experiments}.

\section{Gradient Masking during Attack Optimization: A Case Study }
\label{sec:gradient_obsfucation_or_not}
It is well-known that gradient masking \cite{athalye2018obfuscated, tramer2020adaptive} can cause optimization difficulties during adversary generation, resulting in an inflated robustness. Pang \etal \cite{pang2021bag} show that only moderate use of label smoothing (Eq. \ref{eq:smooth_labels}) helps in boosting adversarial robustness but excessive label smoothing can decrease the robustness. We note that recent state-of-the-art adversarial training methods \cite{feature_scatter, lee2020adversarial} use higher label smoothing during training.  So an important question is: \emph{When does label smoothing causes gradient masking?} Here, we analyse the effect of label smoothing on adversarial robustness as a case study, highlighting the need to diagnose such behaviours \cite{shafahi2019label, fu2020label, lee2020gradient}.

\noindent
\textbf{Proposition 1:} \emph{Consider a model trained using a regular Cross-entropy loss ($\ell_{ce}$) with label smoothing outputs logits $\bm{a}\in \mathbb{R}^{N}$. Then, the gradients used to update the model are relatively weaker i.e., $\partial \ell_{ce}(\bm{a},\widehat{\bm{y}})/ \partial a_i < \partial \ell_{ce}(\bm{a},{\bm{y}})/ \partial a_i$ for $\delta \in (0,1]$ where $\bm{y}\in \mathbb{R}^{N}$ denote the one-hot encoded labels and $\widehat{\bm{y}}$ denotes its smoothed version. As a result, smooth loss surfaces are learned close to the training data manifold, thereby suppressing gradients used to craft adversaries in local neighborhoods. This phenomenon causes gradient masking and leads to inflated  robustness of the learned model in small neighbourhood of the training data \cite{papernot2017practical}.} 

\noindent\textbf{\emph{Sketch Proof:}} For a model being trained with cross-entropy loss $\ell_{ce}$ and one-hot encoded ground-truth labels $\bm{y}$, the gradients are given by,
\begin{align}
    \frac{\partial \ell_{ce}(\bm{a},{\bm{y}})}{ \partial a_i} & = \sigma(a_i) - y_i,\\
    \text{where }  \sigma(a_i) & = \frac{\exp(a_i)}{ \sum_j \exp(a_j)} .
\end{align}
In comparison, the gradients for the same model trained with smoothed labels $\widehat{\bm{y}}$ are given by,
\begin{align}
    \frac{\partial \ell_{ce}(\bm{a},\widehat{\bm{y}})}{ \partial a_i} & = \sigma(a_i) - \widehat{y}_i,\\
    \text{where }\; \widehat{y}_i & = 
    \begin{cases}
    (1- \delta) & \text{if } y_i = 1 \\
    \frac{\delta}{N-1} & \text{if } y_i = 0 
    \end{cases}
\end{align}
where, $N$ is the total number of classes. The above expression shows if $\delta \in (0,1]$, then whether $y_i = 0$ or $y_i =1$, the difference with predicted probability score for a class will always be less than the case when non-smooth labels are used. Hence, 
\begin{align}
     \frac{\partial \ell_{ce}(\bm{a},{\bm{y}})}{ \partial a_i} >  \frac{\partial \ell_{ce}(\bm{a},\widehat{\bm{y}})}{ \partial a_i}.
\end{align}
Label smoothing leads to smooth loss surfaces during model training. However, the existence of large contagious regions of  adversarial pockets in the data manifold \cite{tabacof2016exploring,goodfellow2014explaining} means that such smoothness is only achieved close to the training data manifold. Thus, the gradient directions computed from the same model even for the case of whitebox adversaries, don't disclose directions to which the model still remains susceptible.

\noindent\textbf{Case Study:} We design an adversarial training (AT) algorithm to showcase how an attack fails in the presence of gradient masking. This masking is introduced by label smoothing and we call the resulting AT algorithm as Mask-AT. Mask-AT is based on single step adversarial attack known as FGSM \cite{goodfellow2014explaining} combined with larger random initialization and label smoothing. Our adversarial training (Algo.~\ref{alg:masking_at}) takes large random step in the input space by adding a uniform noise to a given sample and then adversarial examples are computed by taking a single adversarial step using FGSM. Model parameters are updated by minimizing the empirical loss (cross-entropy) with smooth labels on these adversarial examples. We observe that such simple adversarial training shows better robustness against well studied iterative attacks such as PGD while maintaining a high clean accuracy (Fig. \ref{fig:analysis_mask_at}). This behavior resonates well with the recent state-of-the-art defenses including feature scattering (FS) \cite{feature_scatter} and adversarial mixup (AvMix) training \cite{lee2020adversarial}. Thus, our experiment sheds a light on how gradient masking introduced by label smoothing can play a significant part in achieving a high adversarial robustness. 

\input{plots/analysis_on_gradient_masking}

\noindent\textbf{Mask-AT Training:} To study the effect of label smoothing, We train a ResNet18 on CIFAR10 using Mask-AT (Algo. \ref{alg:masking_at}). Models are trained using SGD optimizer for $200$ epochs with batch-size $60$. Pixel values are scaled to $[-1,+1]$. Learning rate is set to $0.1$ and decreased by a factor of $10$ at epochs $60$ and $90$. Perturbation budget, $\epsilon$, is set to ${8}/{255}$ during training. Label smoothing \cite{feature_scatter} is performed as,
\begin{align}
\label{eq:smooth_labels}
    \widehat{\bm{y}} = \big(1 -\delta -\frac{\delta}{N-1}\big)*\bm{y} + \frac{\delta}{N-1},
\end{align}
where $\delta$ is label smoothing factor, $N$ represents number of classes and $\bm{y} \in \mathbb{R}^{N}$ is the one-hot encoded label. Applying Eq.~\ref{eq:smooth_labels} to $\bm{y}$ reduces the confidence of true class to $1-\delta$. 

\noindent\textbf{Mask-AT Evaluation:} We evaluate ResNet18's robustness against PGD attack \cite{madry2018towards}  with 20 iterations and step of $\frac{2}{255}$.

\begin{algorithm}[t]
\small
\caption{Mask-AT}
\label{alg:masking_at}
\begin{algorithmic}[1]
\State A batch of benign samples $\{\bm{x}_{i}, \bm{y}_{i}\}_{i=1}^n$, a model $f$ paramerterized by $\theta$, perturbation budget $\epsilon$, scaling parameter $\eta$, label smoothing factor $\delta$ and cross-entropy loss $\ell_{ce}$.
\For {$i=1$ to $n$} 
\State $\bm{\tilde{x}}_{i} = \bm{x}_{i} + \bm{\mu} \left(\eta\cdot\epsilon \right) : \,\bm{\mu} \sim \text{Uniform}(-1, 1)$
\algorithmiccomment{Take a random step}
\State $\bm{\tilde{x}}_{i}= \text{clip}\left(\textrm{FGSM}(\bm{\tilde{x}}_{i}), \bm{x}_{i}-\epsilon, \bm{x}_{i}+\epsilon \right)$ \algorithmiccomment{Generate adversary}
\EndFor
\State $\theta = \theta - \nabla \ell_{ce}(f(\bm{\tilde{x}}),  \widehat{\bm{y}}(\delta))$ \algorithmiccomment{Update the model parameters}
\end{algorithmic}
\end{algorithm}

\noindent\textbf{Analysis and Observations:}  We dissect each component of Mask-AT (Algo. \ref{alg:masking_at}) in order to better understand the role of label smoothing, random initialization and attack iterations. Results presented in Fig.~\ref{fig:analysis_mask_at} can be analysed as:

$-$\emph{The Effect of Random Initialization:} We set $\delta=0$  i.e.,  no label smoothing is applied.  We perturb the input sample with uniform noise before running the FGSM attack during training (Algo. \ref{alg:masking_at}). Strength of uniform noise is controlled by $\eta$. The higher the $\eta$, the larger the random step taken before the attack. We observe in Fig.~\ref{fig:analysis_mask_at}(a) that model clean accuracy increases by increasing $\eta$ but at the same time its adversarial robustness decreases.  This signifies that training a model with no label smoothing on adversaries computed using FGSM does not make the model robust and it also does not introduce any gradient masking as PGD \cite{madry2018towards} successfully exposes weak robustness. This behavior, however, changes as we introduce label smoothing during training.
    
$-$\emph{The Effect of Label Smoothing:} We now fix the the value of $\eta$ in the next experiment as shown in Fig. \ref{fig:analysis_mask_at}(b-c) and start increasing the label smoothing factor, $\delta$ (Eq. \ref{eq:smooth_labels}). We observe that higher the value of $\delta$, the higher the robustness of the model against PGD attack.  This is an abnormal behavior such that the robustness of ResNet18 trained using Algo. \ref{alg:masking_at} at $\eta=6$ goes from 0\% to 75\% just by increasing label smoothing, $\delta$.
    
$-$\emph{The Effect of Attack Iterations:} Finally, we study if such gradient masking effect continues with iterative attack training as well. We fix the value of $\eta$ and train the model by running 10 attack iterations rather than on FGSM. This is equivalent to Madry \etal \cite{madry2018towards}. We observe that the masking effect caused by the combination of label smoothing and random noise reduces significantly. This indicates that such masking phenomena prevails in models trained with label smoothing and single-step attack such as  \cite{feature_scatter, lee2020adversarial}. 

\noindent\textbf{Outcome and Motivation for Guided Optimization:} Our experimental analysis (Fig. \ref{fig:analysis_mask_at} (a-d)) shows interesting insights about gradient masking. We observe that the same attack (PGD) works perfectly well when there is no gradient masking effect. For example, PGD works as expected when models are trained using FGSM with no label smoothing or Madry's method (with or without label smoothing) but it fails on models trained using FGSM combined with large random initialization and label smoothing. Such uncertainty in attack optimization leads to elusive robustness \cite{feature_scatter, lee2020adversarial, Pang2020Mixup} and there is a need for adversarial attack that performs consistently. 
We provide a complimentary approach to the previous attack solutions \cite{athalye2018obfuscated, croce2020reliable} by proposing to guide optimization using gradient direction from a surrogate model thus avoiding local minima due to gradient masking.


\section{Guided Projected Gradient Attack}
\label{sec:methodology}
In this section, we develop a guiding mechanism based on a new \emph{`match and deceive'} loss that can quickly expose gradient masking with small number of attack iterations. This leads us to a novel attack  named as \emph{Guided Projected Gradient Attack} (G-PGA) attack. (Algo.~\ref{alg:guided_attack}).

\begin{figure*}[t]
    \centering
    \includegraphics[width=0.88\textwidth]{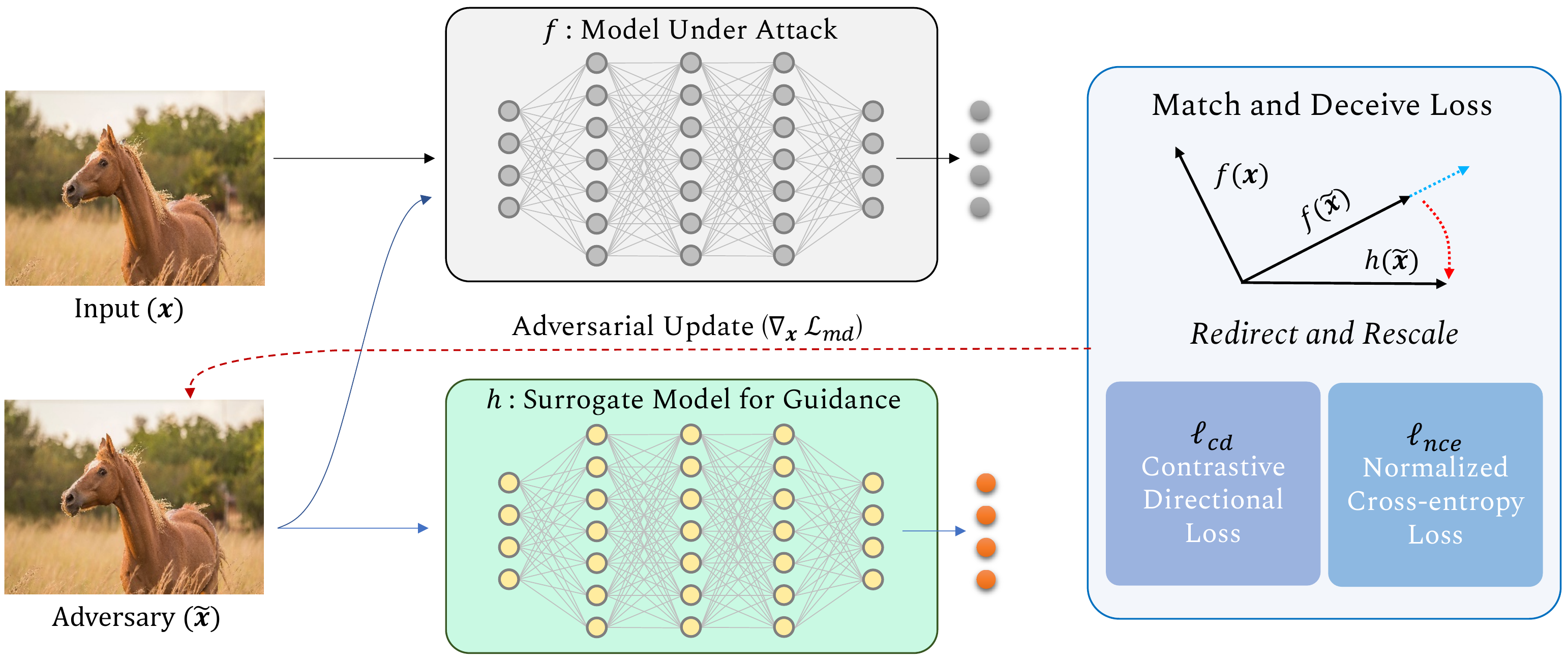}
    \caption{Guided projected gradient attack is based on match and deceive loss which uses information provided by the surrogate model to redirect and attentively rescale the logits of the under attack model in order to overcome gradient masking and find optimal adversarial direction.}
    \label{fig:concept}
\end{figure*}

\subsection{Need for External Guidance}
A \emph{white-box adversary} is created using the full knowledge of model architecture, pretrained weight parameters, training loss along with any randomness used to perturb the input samples during training. Consider a model $f$ parameterized by $\theta$ is adversarially trained using a given training mechanism, $\texttt{train}_{f}$ (e.g., \cite{feature_scatter, lee2020adversarial}). If $\texttt{train}_{f}$ leads to gradient masking (as shown in Sec.~\ref{sec:gradient_obsfucation_or_not}) then white-box adversaries will be less effective against the trained model. 

We observe that the guidance from the loss function can play a key role in launching a successful attack. For example, the adversary is more effective when computed using cross-entropy (CE) loss for over-confident models but becomes less effect on models trained using aggressive label smoothing (Sec.~\ref{sec:gradient_obsfucation_or_not}). Since the guidance available in the white-box settings (with gradient masking) is non-conducive to finding reliable attack directions, the attack needs to look `\emph{elsewhere}' for a better guidance.  To this end, we propose to introduce a surrogate model in the attack pipeline which is used to find optimal attack directions on the source network. 

A naive strategy would be to use a surrogate model, $h$, parameterized by $\phi$ trained using a masking-free method ($\texttt{train}_{h}$) (\eg naturally training using cross-entropy or adversarial training using \cite{madry2018towards, Zhang2019theoretically})  with a similar architecture as model $f$. Adversaries can be computed against model $h$, and then transferred to the model $f$. The problem with this approach is that these adversaries contain gradient noise specific to the model parameters $\phi$ trained using a given approach ($\texttt{train}_{h}$) leading to sub-optimal results for the model $f$ (see Sec.~\ref{sec:surrogate_model} and Fig. \ref{fig:analysis_match_and_deceive} for analytical insights). Note that the adversaries created using model, $h$, are considered \emph{black-box} to the model $f$, if $\texttt{train}_{f} \neq \texttt{train}_{h}$ even when $f$ and $h$ share similar architecture.

\subsection{Match and Deceive Loss} 
Our proposal is to use masking-free information from the surrogate model, $h$, as a guide to find optimal adversarial direction for the model $f$. The objective is to avoid the white-box setting with potential optimization difficulties and the black-box scenario, as both lead to  weaker adversaries. We achieve this guiding mechanism (Fig. \ref{fig:concept}) as follows:
\begin{itemize}\setlength{\itemsep}{0em}
    \item \emph{Redirection:} During the attack, redirect the optimizer with a supervisory signal that matches the adversarial directions from the output space of model $f$, with the surrogate model $h$. Here, the intuition is that if $h$ is masking-free then the optimizer should move along the adversarial direction defined by $h$.  Redirection is achieved using a contrastive directional loss (Eq. \ref{eq:contrastive_directional_loss}).
    \item \emph{Rescaling:} Attentively rescale the logit space outputs of the under-attack model using the guidance of logits from surrogate model. 
    Rescaling allows us to incorporate masking free information from the surrogate into the cross-entropy loss. This is achieved by a normalized cross-entropy loss (Eq. \ref{eq:logit_loss}). 
\end{itemize}

\begin{algorithm}[t]
\small
\caption{Guided Projected Gradient Attack}
\label{alg:guided_attack}
\begin{algorithmic}[1]

\State A benign sample $\bm{x}$, a classifier $f$, a surrogate model $h$, perturbation budget $\epsilon$, number of attack iterations $T$, step size $\kappa$.

\State $\bm{g}_0 = 0$; $\bm{\tilde{x}} = \bm{x}$, $t \leftarrow  0$;
\Repeat
\State $t \leftarrow  t + 1$;
\State Forward pass  $\bm{x}, \bm{\tilde{x}}$ through $f$ and $h$ and compute $f(\bm{x})$, $f(\bm{\tilde{x}})$ and   $h(\bm{\tilde{x}})$.
\State Compute contrastive directional loss $\ell_{cd}$ (Eq. \ref{eq:contrastive_directional_loss}) using $f(x)$, $f(\bm{\tilde{x}})$ and $h(\bm{\tilde{x}})$.
\State Compute normalized cross-entropy loss $\ell_{nce}$ (Eq. \ref{eq:logit_loss}) using $f(\bm{\tilde{x}})$ and $h(\bm{\tilde{x}})$.
\State Compute the match and deceive loss $\mathcal{L}_{md}$ (Eq. \ref{eq:md}).
\State Backward pass and compute gradients $\bm{g}_{t} = \nabla_{\bm{x}}\mathcal{L}_{md}$.
\State Use gradients to update perturbation estimate,
$$
\bm{\tilde{x}}_{t+1} = \bm{\tilde{x}}_{t} +\kappa\cdot\mathrm{sign}(\bm{g}_{t}). \notag
$$
\State Project the adversary within allowed perturbation budget, $\epsilon$
\begin{equation}
\bm{\tilde{x}}_{t+1} = \mathrm{clip}(\bm{\tilde{x}}_{t+1}, \hspace{1ex} \bm{x}-\epsilon,  \hspace{1ex} \bm{x}+\epsilon). \notag
\end{equation}
\Until{$t \le T$}
\end{algorithmic}
\end{algorithm}

\noindent\textbf{Contrastive Directional Loss:} Consider a benign input sample $\bm{x}$, an adversarially perturbed sample $\bm{\tilde{x}}$, 
then contrastive directional loss $\ell_{cd}$ minimizes the similarity between the output vectors $f(\bm{x}) \in 
\mathbb{R}^k$ and $f(\bm{\tilde{x}}) \in  \mathbb{R}^k$ in the $k$-dimensional logit-space, a desirable property for an optimal adversary to maximally perturb the input.  It simultaneously maximizes the similarity between $f(\bm{\tilde{x}})$ and $h(\bm{\tilde{x}}) \in  \mathbb{R}^k$ which provides better adversarial gradient direction considering $h$ is adversarially trained and masking free. The $\ell_{cd}$ loss is defined as,
\begin{align}
\label{eq:contrastive_directional_loss}
    \ell_{cd} = -\log \frac{\exp({\mathcal{S}(f(\bm{\tilde{x}}), h(\bm{\tilde{x}})))}} {\exp({\mathcal{S}(f(\bm{x}), f(\bm{\tilde{x}})))} + \exp({\mathcal{S}(f(\bm{\tilde{x}}), h(\bm{\tilde{x}})))} },
\end{align}
where $\mathcal{S}(\cdot)$ represents the cosine similarity between two given vectors i.e., $\mathcal{S}(\bm{a}, \bm{b}) = \frac{\bm{a}^{\top}\bm{b}}{\| \bm{a}\|\| \bm{b}\|}$.

\noindent \textbf{Normalized Cross-Entropy:} As $f(\bm{\tilde{x}})$ and $h(\bm{\tilde{x}})$ represent the logit response for the perturbed sample $\bm{\tilde{x}}$, then guidance through logit rescaling is defined as: 
\begin{equation}
    \mathrm{I}\big(f(\bm{\tilde{x}}), h(\bm{\tilde{x}})\big) = \frac{f(\bm{\tilde{x}}) \circ  h(\bm{\tilde{x}})}{\|f(\bm{\tilde{x}})\|_{2}}, 
\end{equation} 
where, $\circ$ denotes Hadamard product. Adversarial perturbation can be created by maximizing the following normalized cross-entropy loss:
\begin{equation}
\label{eq:logit_loss}
\ell_{nce} = -\sum_{j=1}^{k} y_{j}\log\big(\sigma(\mathrm{I}(f(\bm{\tilde{x}}),  h(\bm{\tilde{x}}))_{j})\big), 
\end{equation}
where $k$ represents the number of classes, $\sigma$ is a softmax function and $\bm{y} \in \mathbb{R}^k$ is the corresponding one-hot encoded ground-truth vector. The $\ell_{nce}$ loss promotes both $f(\bm{\tilde{x}}$) and $h(\bm{\tilde{x}})$ to agree on the misclassification of the perturbed sample with a similar confidence.

The final match and deceive loss is given as follows:
\begin{equation}
\label{eq:md}
    \mathcal{L}_{md} =  \ell_{nce} - \ell_{cd}.
\end{equation}
The above loss is maximized to obtain optimal adversaries in our proposed guided attack.

\subsection{Gradient Analysis}\label{sec:grad_anal}
For the sake of brevity, we consider $\bm{u} = f(\tilde{\bm{x}})$, $\bm{v} = f(\bm{x})$ and $\bm{z} = h(\tilde{\bm{x}})$. Let's assume their corresponding normalized versions are denoted as $\bm{u}' = \bm{u}/\|\bm{u} \|$, $\bm{v}' = \bm{v}/\|\bm{v} \|$ and $\bm{z}' = \bm{z}/\|\bm{z} \|$. Then the loss function $\ell_{cd}$ can be written in terms of dot-product as follows:
\begin{align}
    \ell_{cd} = - \log \frac{ \exp(\bm{u}' \cdot \bm{z}')}{\exp(\bm{v}'\cdot \bm{u}') + \exp(\bm{u}' \cdot \bm{z}') }.
\end{align}
In the following, we compute the gradients of contrastive directional loss function. Since, our goal is to learn the adversarial image $\tilde{\bm{x}}$, we are mainly interested in the gradients  $\frac{\partial \ell_{cd}}{\partial \bm{u}}$ and $\frac{\partial \ell_{cd}}{\partial \bm{z}}$ involving the base model and surrogate model, respectively. These gradients are given by (proof in Appendix), 
\begin{align}
    \frac{\partial \ell_{cd}}{\partial \bm{u}} =  - \frac{ (\bm{z}'- \bm{v}') \exp(\bm{v}'\cdot \bm{u}') (\bm{I} - \bm{u}' \cdot \bm{u}'^{\top})}{ \| \bm{u}\| (\exp(\bm{v}'\cdot \bm{u}') + \exp(\bm{u}' \cdot \bm{z}') )  }, 
    \\
    \frac{\partial \ell_{cd}}{\partial \bm{z}} = - \frac{ \bm{u}'   \exp(\bm{v}'\cdot \bm{u}') (\bm{I} - \bm{z}' \cdot \bm{z}'^{\top}) }{ \| \bm{z}\|(\exp(\bm{v}'\cdot \bm{u}') + \exp(\bm{u}' \cdot \bm{z}') ) }.
\end{align}

\begin{figure}[!t]
\centering
  \begin{minipage}{.48\linewidth}
  	\centering
    \includegraphics[ width=\linewidth,keepaspectratio]{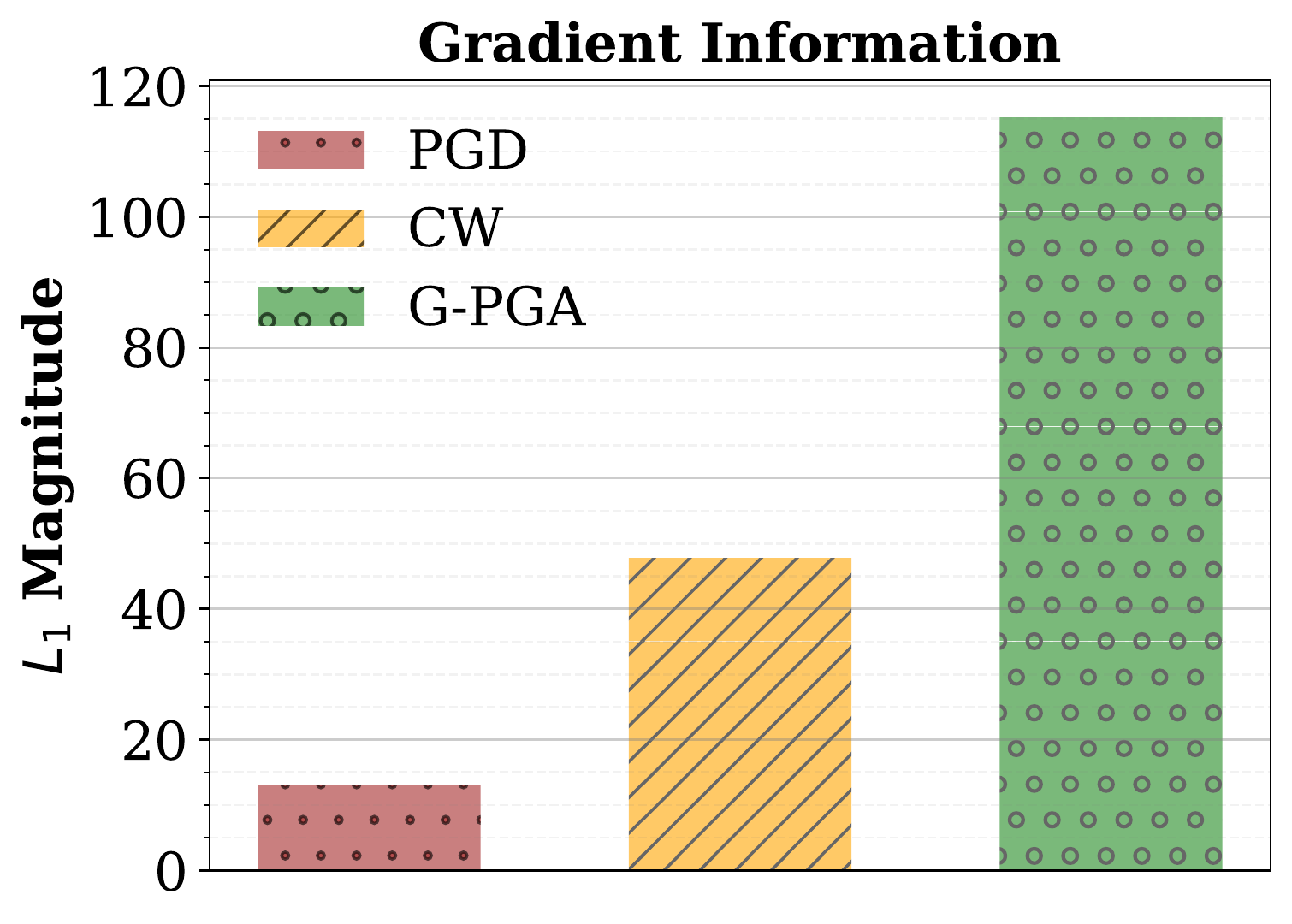}
  \end{minipage}
  \begin{minipage}{.48\linewidth}
  	\centering
    \includegraphics[width=\linewidth, keepaspectratio]{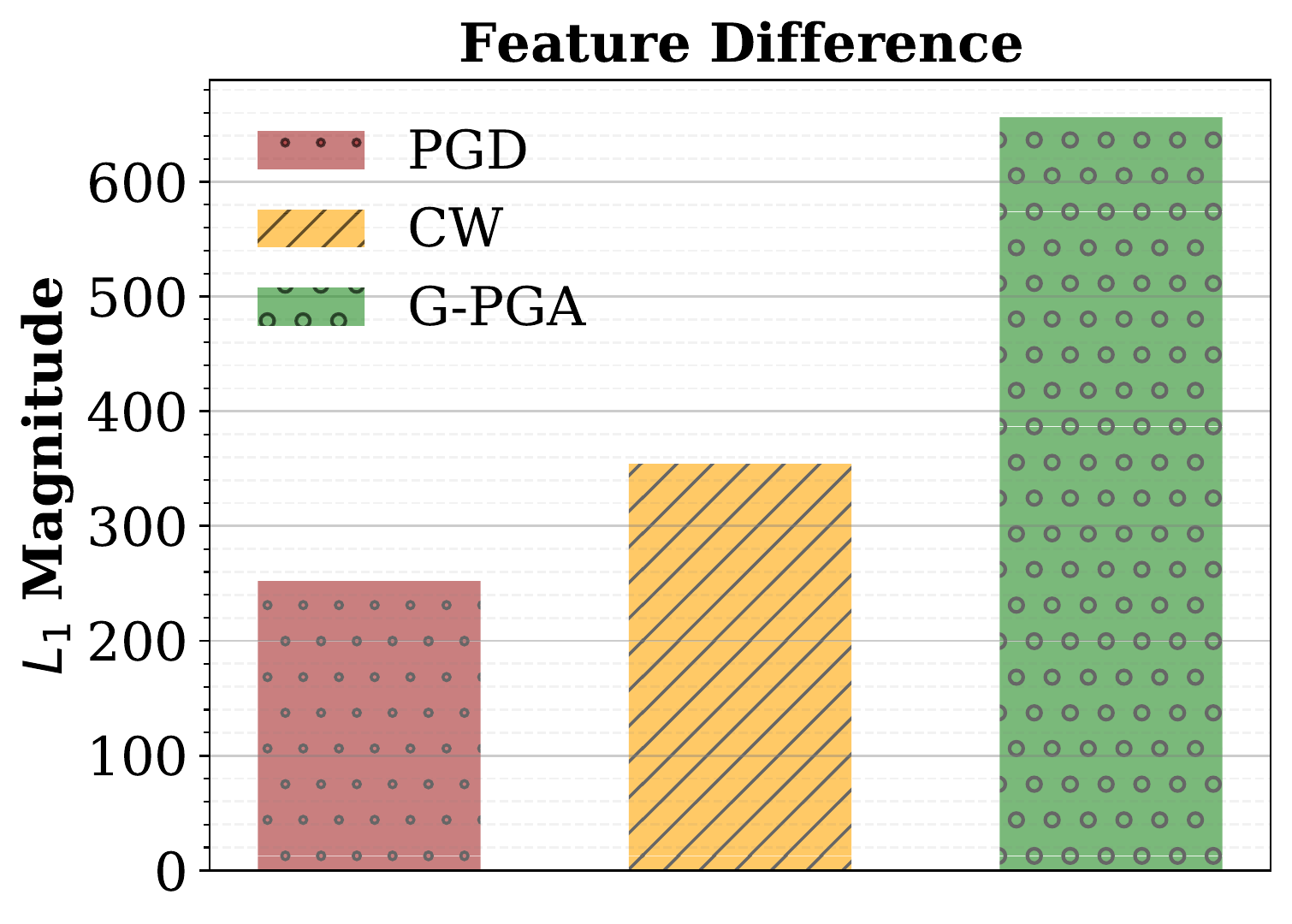}
  \end{minipage}
  \caption{$L_{1}$ magnitude averaged over 10k CIFAR10 test samples (\emph{higher is better}). \emph{Left:} our attack increases the gradient information vital for creating effective adversaries. \emph{Right:} $L_1$ distance (higher the better) between adversarial and clean features (extracted before the logit layer).}
\label{fig:gradiet_and_feature_analysis}
\end{figure}

Fig.~\ref{fig:gradiet_and_feature_analysis} shows empirical analysis for the  gradient information provided by the match and deceive loss. We run PGD, CW and G-PGA (our) attack for 20 iterations against Mask-AT model and compare the average magnitude of gradients across attacks. We also show the feature distortion caused by the three attacks on CIFAR10. Our results show that the proposed loss can provide stronger gradients with guidance from the surrogate model.


\section{Evaluations}\label{sec:experiments}
\subsection{Experimental Protocols}
We evaluate the effect of guided attack optimization on different adversarial training methods including Madry \etal \cite{madry2018towards}, Trades \cite{Zhang2019theoretically}, Feature Scattering (FS) \cite{feature_scatter}, AvMix \cite{lee2020adversarial} and Mask-AT (Algo. \ref{alg:masking_at}). Pretrained model (CIFAR10) for FS is publicly available. Rest of the models are reproduced using open source code bases \cite{feature_scatter, lee2020adversarial, Zhang2019theoretically}. We used open source implementation of attacks including Auto-Attack \cite{croce2020reliable}, FAB \cite{croce2020minimally}, and Square attack \cite{andriushchenko2019square}.  Training and evaluation are performed on four commonly used datasets: CIFAR10, CIFAR100, SVHN, and ImageNet and we report Top-1 test set accuracy (\%). All experiments are conducted using using Nvidia Tesla-V100 with Pytorch library \cite{paszke2019pytorch}. The adversarial perturbations are $l_{\infty}$ bounded and clearly mentioned in each experiment.  
We now evaluate our proposed attack under two settings: \textbf{(a)} \emph{Standalone Efficiency of G-PGA}, \textbf{(b)} \emph{Effect of G-PGA in Ensemble of Attacks}, and \textbf{(c)} \emph{Effect of Guidance to Noisy Gradient Estimation}. 

\input{tables/main_table}

\begin{table*}[!t]
	\centering\small
		\setlength{\tabcolsep}{3pt}
		\scalebox{0.9}[0.9]{
		\begin{tabular}{lccccccccc}
				\toprule
				\rowcolor{Gray} 
			\multirow{3}{*}{Defense} & & \multicolumn{8}{c}{Effect of our G-PGA on Auto-Attack}\\
			\cmidrule(lr){3-10} 
			&& {CE}  & {CE+Ours}  &  {DLR} &  {DLR+Ours} & Croce \emph{et al.} \cite{croce2020minimally} &\cite{croce2020minimally}{+Ours} & Andriushchenko \emph{et al.} \cite{andriushchenko2019square} &  \cite{andriushchenko2019square}{+Ours} \\
			\cmidrule(lr){3-10} 
			& \texttt{Cost$\rightarrow$}& 100 & 20 & 100 &20 & 100 & 50 & 5k(Q)& 10+1k(Q)\\
			\midrule
			FS (CIFAR10) &&64.2&42.7&48.9&41.7&40.84&38.9&59.12&41.2\\
			FS (CIFAR100)&&44.8&1.01&3.0&0.71&8.34&0.74&24.7&1.0\\
			FS (SVHN)    &&24.5&12.3&20.2&8.4&15.8&7.9&65.7&43.5\\
			\bottomrule
	\end{tabular}}
	\caption{\centering Our proposed attack, G-PGA, significantly enhances attack success rate along with convergence when used in combination of attacks proposed by \cite{croce2020reliable}. These results compliment guided optimization provided by G-PGA. 'Q' represents number of queries sent to the model to estimate gradients. Top-1 (\%) accuracy is reported (\emph{lower is better}).}
	\label{tab:comparison_with_AA}
\end{table*}

\begin{table}[!t]
	\centering
		\setlength{\tabcolsep}{7pt}
		\scalebox{1}[1]{
		\begin{tabular} { ccccc }
			\toprule
			\rowcolor{Gray}
			& &\multicolumn{3}{c}{ResNet50} \\
			\rowcolor{Gray} 
			\multirow{-2}{*}{Dataset}&\multirow{-2}{*}{Attack} & $\epsilon=2.0$& $\epsilon=4.0$& $\epsilon=8.0$\\
			 \hline
			 \multirow{3}{*}{ImageNet-5k} &PGD  & 67.6 & 50.7  &32.1\\
			 & Auto-Attack  & 65.6 & 47.5&\textbf{30.0}\\
		  & G-PGA & \textbf{64.7} & \textbf{47.3}  &\textbf{30.0}\\
			 \bottomrule
	\end{tabular}}
	\caption{ \emph{Effectiveness of G-PGA on ImageNet:} We evaluate adversarially trained ResNet50 models from \cite{salman2020adversarially} against different attacks including G-PGA (ours). Note that evaluation perturbation budget for each model is the same as its training perturbation budget 
	\eg ResNet50 trained on $\epsilon=2.0$ is evaluated with the perturbation budget of $\epsilon=2.0$. The surrogate model used in G-PGA is a naturally trained ResNet50. G-PGA performs favorably in comparison to Auto-Attack while being computationally efficient (Table \ref{tab:computational_comparison_with_AA}).  }
	\label{tab: imagenet_results}
\end{table}

\begin{table}[!t]
	\centering\small
		\scalebox{0.85}{
		\begin{tabular}{lcccc}
				\toprule
					\rowcolor{Gray}
			Attack & CIFAR10 & CIFAR100 & SVHN & ImageNet-5k\\
			\midrule
            Auto-Attack & 682.5&2598.2&1774.5&12.8$\times$10$^3$\\
            Auto-Attack (Cheap) & 148.5&581.0&388.7&4252 \\ 
            Proposed Attack &\textbf{ 87.0} & \textbf{88.2}&\textbf{226.2}&\textbf{240}\\
			\bottomrule
	\end{tabular}}
	\caption{Computational time (minutes, \emph{lower is better}) is noted on a Tesla-V100. Attacks ran on $l_\infty$ adversarially trained models at $\epsilon=8$. WideResNet is used for CIFAR10, CIFAR100, and SVHN datasets while ResNet50 is used for ImageNet-5k. Results are reported using the test samples for each dataset with a batch size of 100.}
	\label{tab:computational_comparison_with_AA}
\end{table}

\subsection{Standalone Efficiency of G-PGA}
We compare our attack strength with baseline methods including PGD \cite{madry2018towards} and CW. CW attack is based on a margin loss \cite{carlini2017towards} used by FS \cite{feature_scatter} and AvMix \cite{lee2020adversarial} in PGD attack in a restricted perturbation setting. Step size is set to 2/255 for all the attacks. All experiments in Table \ref{tab:main-results} are conducted on WideResNet
. Following insights emerge from our experiments: (a) Our attack perform significantly better than PGD and CW while having 50\% less iterations. It decreases the robustness of FS~\cite{feature_scatter}, AvMix~\cite{lee2020adversarial} and Mask-AT  significantly (< 2\%) as the number of classes increases \eg on CIFAR100 dataset (Table \ref{tab:main-results}), and (b) Madry \etal \cite{madry2018towards} and Trades~\cite{Zhang2019theoretically} are least affected by the masking effect of label smoothing but our attack remains stronger against these defenses.

\begin{figure}[!t]
\centering
  \begin{minipage}{.48\linewidth}
  	\centering
    \includegraphics[ width=\linewidth,keepaspectratio]{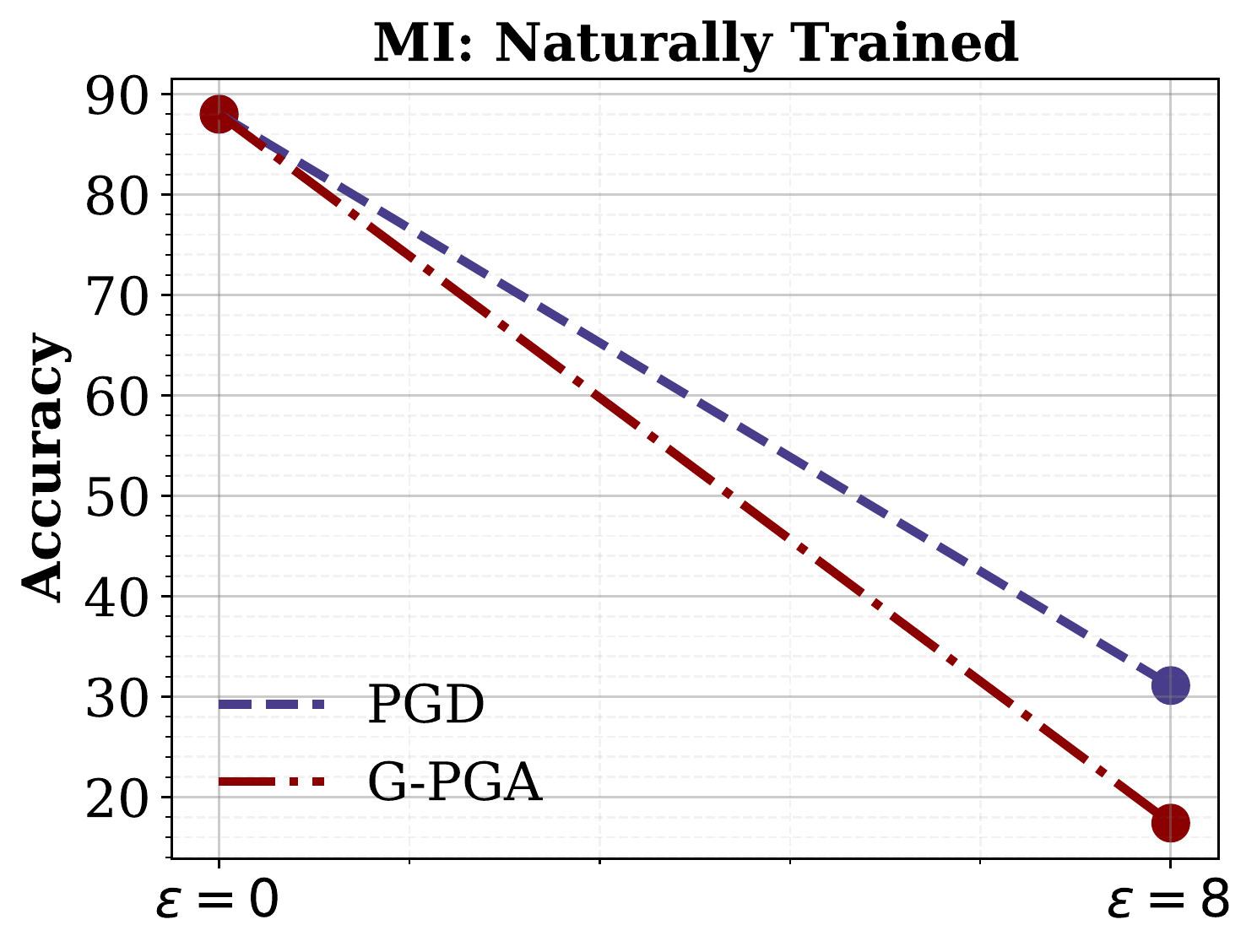}
  \end{minipage}
  \begin{minipage}{.48\linewidth}
  	\centering
    \includegraphics[width=\linewidth, keepaspectratio]{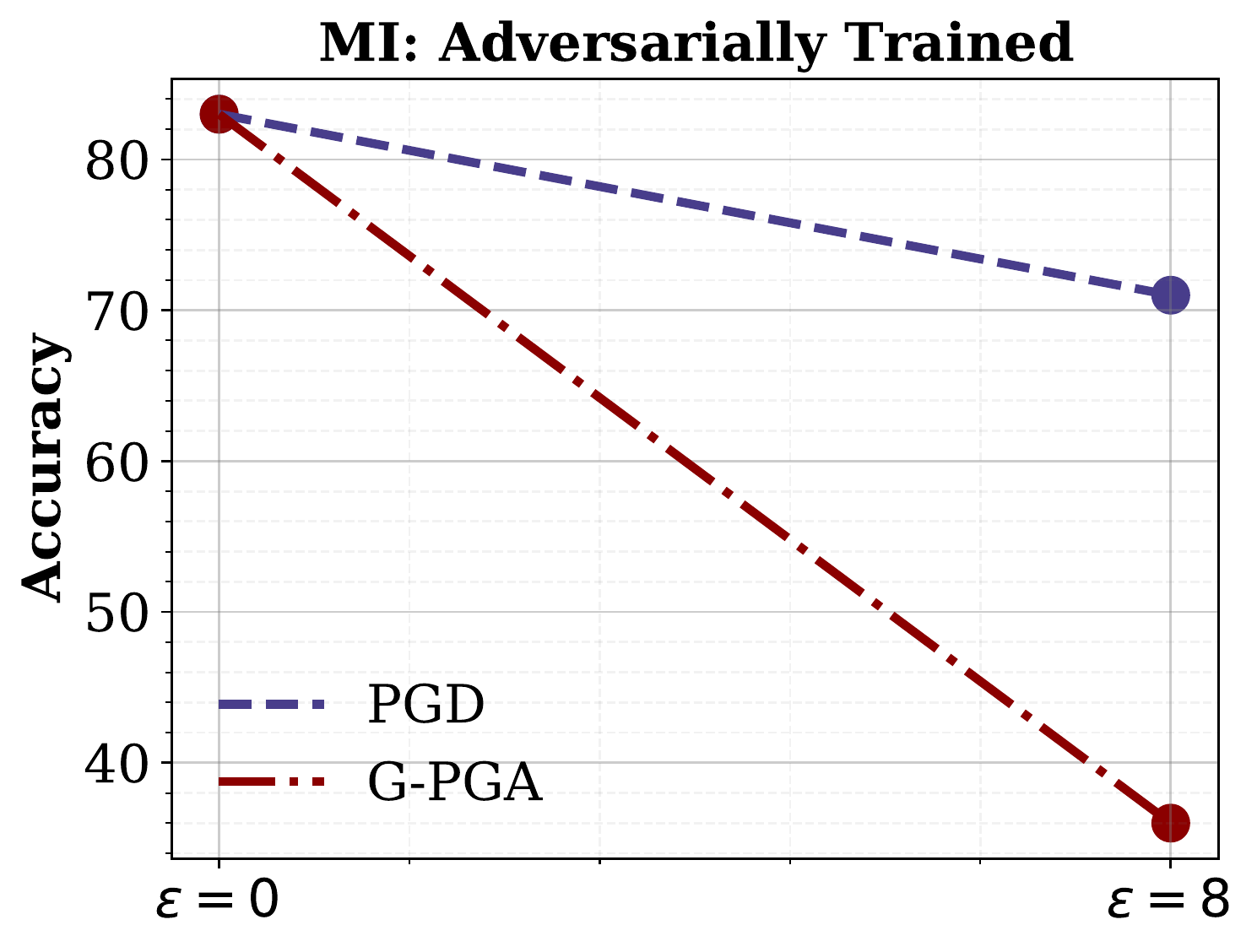}
  \end{minipage}
    \caption{Perturbations found using our attack break MI \cite{Pang2020Mixup} without adapting to the defense. Top-1 (\%) accuracy is reported on 1k CIFAR10 test samples with WideResNet (\emph{lower is better}). MI is performed with a combination of predicted and other labels as in \cite{Pang_github}.  }
  \label{fig:breaking_mi}
\end{figure}
	
\noindent\emph{Breaking Mixup-Inference (MI):} Mixup \cite{zhang2017mixup} is another way to smooth model output predictions. Pang \etal \cite{Pang2020Mixup} proposed to break the locality of adversarial examples by exploiting global linear behavior of the model after Mixup training (natural or adversarial). They propose to mix adversarial sample before inference to reduce attack strength. However, our proposed perturbations (Algo. \ref{alg:guided_attack}) significantly reduces the effect of such dynamic inference which reflects our attack's strength (see Fig. \ref{fig:breaking_mi}). It is important to note that the attack is not adapted specifically to MI defense \ie we are able to break the model without any knowledge about the Mixup inference defense.

\subsection{Effect of G-PGA in Ensemble of Attacks} Croce \etal \cite{croce2020reliable} deploy multiple attacks including modified versions of PGD \cite{madry2018towards}, FAB \cite{croce2020minimally} and query-based Square \cite{andriushchenko2019square} attack. Our proposed attack is computationally less expensive \eg 85\% less costly to run on CIFAR100 in comparison to the cheaper version of AutoAttack \cite{croce2020reliable} (see Table \ref{tab:computational_comparison_with_AA}) but it performs on par to \cite{croce2020reliable} without the need for large number of iterations, random restarts or thousands of queries. However, the guiding mechanism we presented is generic in its nature. Therefore, it can be used in combination of other attacks as proposed by \cite{croce2020reliable}. Hence, we study the effect of G-PGA in combination to each of the four untargeted attacks proposed by \cite{croce2020reliable} including PGD variants based on CE and DLR loses \cite{croce2020reliable}, FAB \cite{croce2020minimally} and query based Square attack \cite{andriushchenko2019square}. We observe in Table \ref{tab:comparison_with_AA} that our guided mechanism enhances the efficiency of each component of AutoAttack while decreasing their computational cost. To highlight an example, when combined with G-PGA, square attack decreases accuracy of model trained using \cite{feature_scatter} on CIFAR100 from 24\% to 1\% within only 1k queries.

\begin{figure*}[!t]
\centering
  \begin{minipage}{.325\textwidth}
  	\centering
    \includegraphics[ width=\linewidth,keepaspectratio, clip=true,trim=0 18mm 0 0mm]{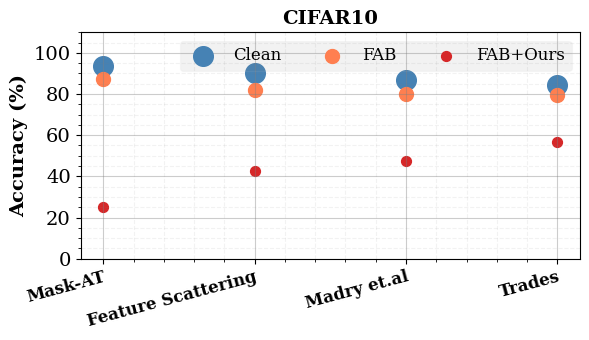}
  \end{minipage}
  \begin{minipage}{.325\textwidth}
  	\centering
    \includegraphics[width=\linewidth, keepaspectratio, clip=true,trim=0 18mm 0 0mm]{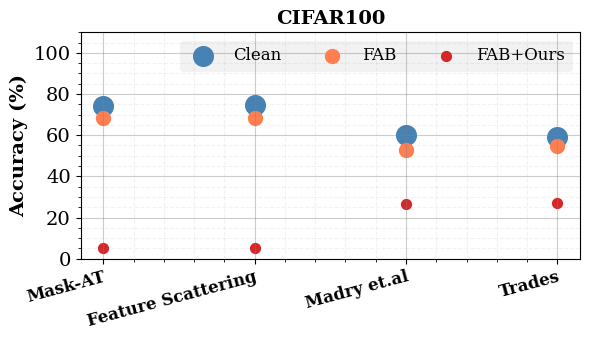}
  \end{minipage}
  \begin{minipage}{.325\textwidth}
  	\centering
    \includegraphics[width=\linewidth, keepaspectratio, clip=true,trim=0 18mm 0 0mm]{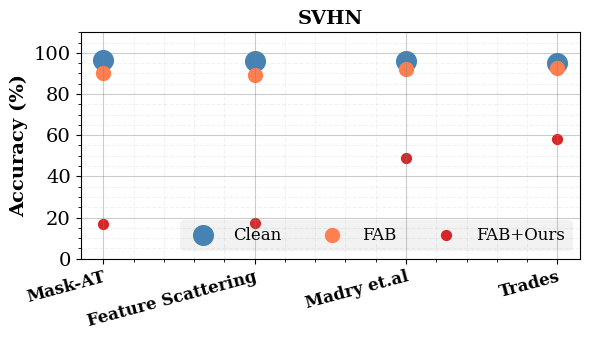}\
  \end{minipage}
  
    \begin{minipage}{.325\textwidth}
  	\centering
    \includegraphics[ width=\linewidth,keepaspectratio, clip=true,trim=0 0 0 0mm]{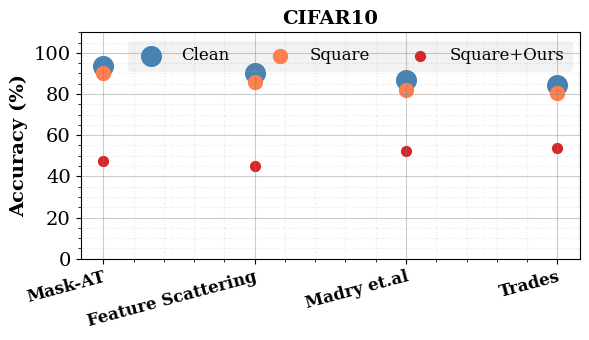}
  \end{minipage}
  \begin{minipage}{.325\textwidth}
  	\centering
    \includegraphics[width=\linewidth, keepaspectratio, clip=true,trim=0 0 0 0mm]{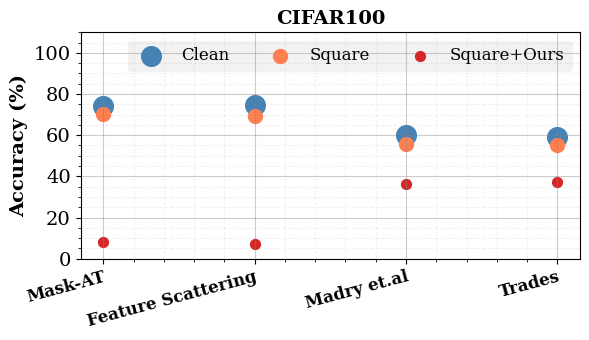}
  \end{minipage}
  \begin{minipage}{.325\textwidth}
  	\centering
    \includegraphics[width=\linewidth, keepaspectratio, clip=true,trim=0 0 0 0mm]{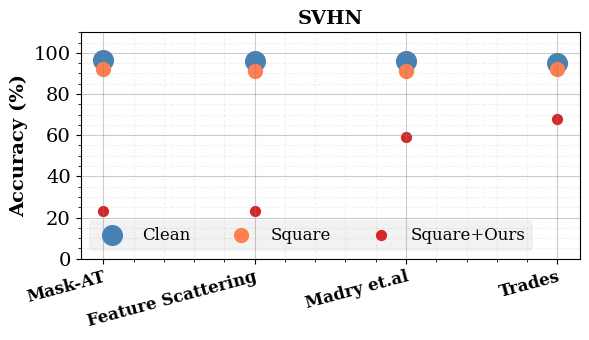}\
  \end{minipage}
  \caption{\emph{Guidance to Noisy Gradient Estimation:} Boundary based attack FAB \cite{croce2020minimally} (top row) and query based Square attack \cite{andriushchenko2019square} (bottom row) failed to estimate adversarial direction when adversarially trained models are protected by a dynamic defense. The original (under attacked) model perturbs the  incoming input sample $\bm{x}$ (clean or adversarial) with uniform noise ($\bm{x} + \bm{\mu} \left(\eta\cdot\epsilon \right), \text{ where } \eta=1, \epsilon=8 $) during inference. We use 5000 queries for Square attack and FAB attack ran for 100 iterations. Note that under attacked defenses are trained on adversarial examples with the same perturbation budget of $\epsilon=8$. Therefore such dynamic defense causes minimal drop in clean accuracy, however,  theses attacks failed mainly because of noisy and sub-optimal gradient estimation.  Our guiding mechanism allows the same attacks \cite{croce2020minimally, andriushchenko2019square} (top and bottom row respectively) to estimate the optimal adversarial direction with the help of surrogate model and quickly breaks the original model. Top-1 (\%) accuracy is reported on the test sets of each dataset (\emph{lower is better}). Results are the average of five runs.} 
\label{fig:defenses_vs_GPGA}
\end{figure*}

\subsection{Effect of G-PGA against no Label Smoothing}
Here, we evaluate different defenses that are not dependent on label smoothing during training. These defense methods such as Madry \etal \cite{madry2018towards} and Trades \cite{Zhang2019theoretically} do not suffer from masking so the performance simple PGD \cite{madry2018towards}, auto-attack \cite{croce2020reliable} and G-PGA is equivalent. Results are presented in Table \ref{tab: evaluation_at_delta_0}. We note simple attack such as PGD \cite{madry2018towards} are effective against defenses that do not suffer from masking effect, however, our guided attack can expose true robustness even when model suffer from gradient masking effect thus leading to more reliable robust evaluation. This further highlights that when large label smoothing ($\delta=0.5$) is used with iterative adversarial training \cite{madry2018towards, Zhang2019theoretically}, it does not introduce gradient masking but can reduce the adversarial robustness of a model. As an example, adversarial robustness of Trades \cite{Zhang2019theoretically}  reduces from 52.8 at $\delta=0.0$ (Table \ref{tab: evaluation_at_delta_0}) to 51.9 at $\delta=0.5$ (Table \ref{tab:main-results}). 

We further validate our approach on the large-scale ImageNet \cite{deng2009imagenet} dataset.   Our approach consistently produces favorable results while being computationally efficient (Tables \ref{tab: imagenet_results} and \ref{tab:computational_comparison_with_AA}). We evaluated publicly available adversarially trained models \cite{salman2020adversarially} on a subset of ImageNet (5k samples) against different attacks. The surrogate model used in our G-PGA attack is simply a naturally trained ResNet50 model which is also publicly available \cite{paszke2019pytorch}. These results show the generalizability of our method across datasets and different surrogate models as well (Fig. \ref{fig:surrogate_effect_of_trainings}).

\subsection{Guidance to Noisy Gradient Estimation}
One strong feature of our proposed guidance is the masking free attack optimization. We empirically validate this hypothesis by observing if our method allows to find useful adversarial directions for query based adversarial attacks when the gradient estimation becomes unreliable from the original model (Fig. \ref{fig:defenses_vs_GPGA}). In this case, attacker has access to only the model's output and needs to estimate the gradients. Auto-Attack \cite{croce2020reliable} also relies on query based Square \cite{andriushchenko2019square} and boundary based FAB \cite{croce2020minimally} attacks to estimate the gradients. However, defending against such attacks \cite{croce2020minimally, andriushchenko2019square} by injecting noise into the input image or the model's output has been motivated and well studied in \cite{byun2022effectiveness, qin2021random}. These defenses  protect the model by corrupting either the input image \cite{byun2022effectiveness} or the model outputs (logits) \cite{qin2021random} with the random noise and thereby corrupting the gradient estimation. Since input and output signal from the model is corrupted at each query during attack optimization, therefore such practical attacks \cite{croce2020minimally, andriushchenko2019square} struggle to adapt to the deployed defense and can not find optimal adversarial directions. 

As demonstrated in Table \ref{tab:comparison_with_AA}, FAB attack can reduce the robustness of feature scattering \cite{feature_scatter} to 40.84\% but fails when feature scattering is further protected with random noise defenses \cite{byun2022effectiveness, qin2021random} (Fig. \ref{fig:defenses_vs_GPGA}). The same observations can be made for the Square attack. These findings are consistent across different datasets (CIFAR10, CIFAR100, and SVHN) and training approaches (Mask-AT (ours), Feature Scattering \cite{feature_scatter}, Madry \etal \cite{madry2018towards}, and Trades \cite{Zhang2019theoretically}) (Fig. \ref{fig:defenses_vs_GPGA}).

Our proposed guidance compliments these attacks to adapt to such dynamic defenses and fools the under attacked model within few queries \cite{andriushchenko2019square} or iterations \cite{croce2020minimally} (Fig. \ref{fig:defenses_vs_GPGA}).  The  results of this new experiment shed light on how the masking free guidance through a surrogate model allows an attack to avoid being stuck in non-optimal solution even when gradient estimation from the under attacked model is sub-optimal.

\input{tables/evalutaion_with_no_label_smoothing}

\begin{figure*}[!t]
\centering
  \begin{minipage}{.245\textwidth}
  	\centering
  	\small Clean Images
    \includegraphics[ width=\linewidth,keepaspectratio, clip=true,trim=0 0mm 0 0mm]{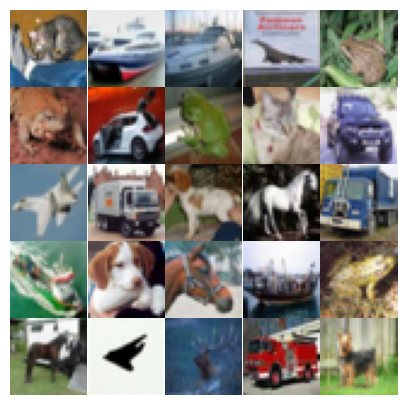}
    
  \end{minipage}
  \begin{minipage}{.245\textwidth}
  	\centering
  	\small No Attack
    \includegraphics[width=\linewidth, keepaspectratio, clip=true,trim=0 0mm 0 0mm]{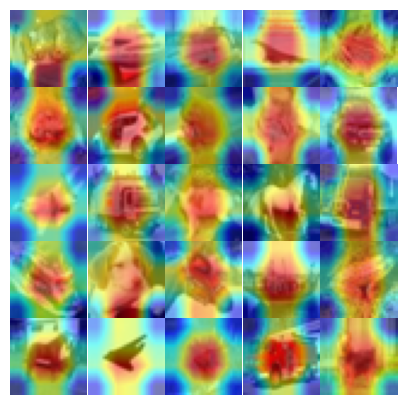}
    
  \end{minipage}
  \begin{minipage}{.245\textwidth}
  	\centering
  	 \small PGD Adversaries
    \includegraphics[width=\linewidth, keepaspectratio, clip=true,trim=0 0mm 0 0mm]{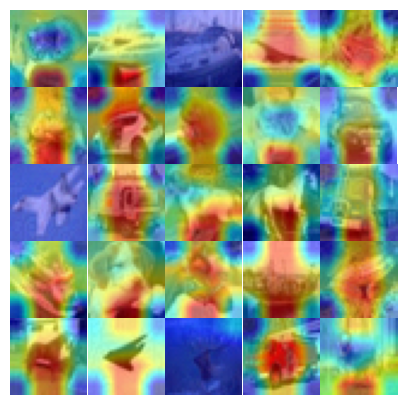}\
  \end{minipage}
    \begin{minipage}{.245\textwidth}
  	\centering
  	G-PGA Adversaries
    \includegraphics[ width=\linewidth,keepaspectratio, clip=true,trim=0 0 0 0mm]{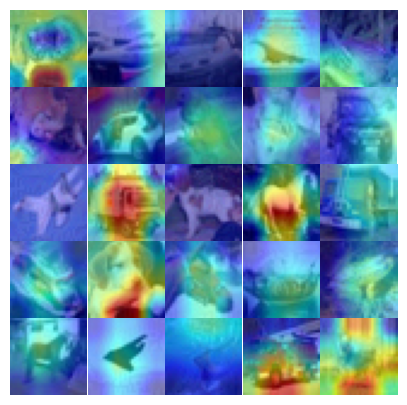}
  \end{minipage}
  \caption{\emph{Attention Visualization of Adversarial Features:} Adversarial examples can be explained by the adversarial features imprinted on the original image within a certain perturbation budget \cite{ilyas2019adversarial, Naseer_2021_ICCV}. We visualize presence of the true class features in the input samples using \cite{selvaraju2017grad}. We observe that adversaries generated by a failed attack (PGD) still contains the features of true class as indicated by the attention maps. G-PGA on the other hand successfully maximizes adversarial features and minimizes the presence of true class features as indicated by the dispersed attention. This validates the effectiveness of our approach. } 
\label{fig:attn_maps}
\end{figure*}

\begin{figure*}[!t]
\centering
  \begin{minipage}{.325\textwidth}
  	\centering
    \includegraphics[ width=\linewidth,keepaspectratio, clip=true,trim=0 0mm 0 0mm]{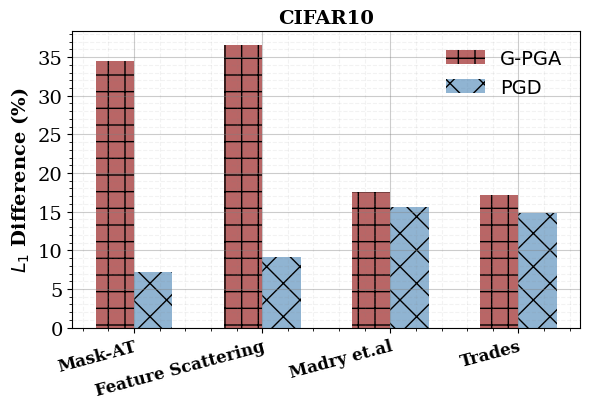}
  \end{minipage}
  \begin{minipage}{.325\textwidth}
  	\centering
    \includegraphics[width=\linewidth, keepaspectratio, clip=true,trim=0 0mm 0 0mm]{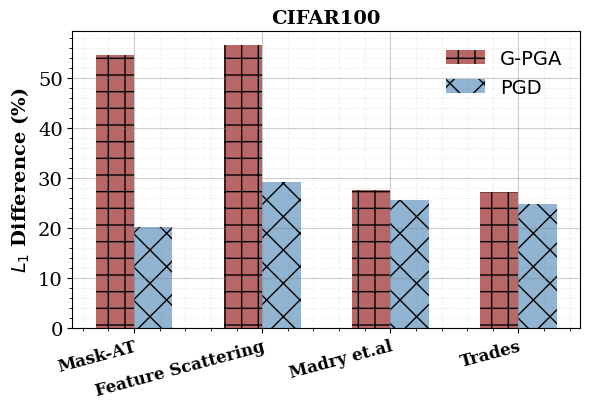}
  \end{minipage}
  \begin{minipage}{.325\textwidth}
  	\centering
    \includegraphics[width=\linewidth, keepaspectratio, clip=true,trim=0 0mm 0 0mm]{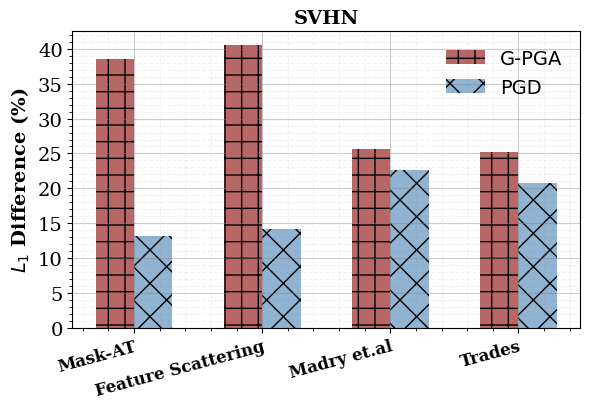}\
  \end{minipage}
  
  \caption{\emph{Quantify Attention Dispersion:} Our attack approach, G-PGA, minimizes the presence of the features of true class within an allowed perturbation budget (Fig. \ref{fig:attn_maps}). We quantify this by measuring the $L_1$ difference (\emph{higher is better}) between attention maps of true class produced by \cite{selvaraju2017grad} between adversarial and the clean images. We observe that in comparison to PGD, G-PGA produces large difference for the models suffering from gradient masking. This highlights that our proposed guidance can successfully reveal gradient masking by minimizing the features of the true class during attack optimization. Results are reported on the test set of each dataset and averaged across the total number of samples. } 
\label{fig:attn_maps_quantity}
\end{figure*}

\input{plots/ablation_whitebox_vs_blackbox_vs_ours}

\section{Ablative Analysis}
\label{sec:surrogate_model}
\subsection{Attention Visualization of Adversarial Features}
Ilyas \etal \cite{ilyas2019adversarial} showed that adversarial examples can be explained by the features of the mis-classified class labels. We visualize the adversarial features produced with and without our proposed guidance by observing the attention maps of the true class using \cite{selvaraju2017grad}. If an attack is stuck in a local minima then its adversarial features will be weak. Fig. \ref{fig:attn_maps} shows attention maps of the adversarial images of a failed attack (PGD) which are closer to the clean images (without any adversarial feature/noise). This means that attack failed to suppress the features of true class during optimization. On the other hand, attention is dispersed on the adversarial images generated using our guided attack (G-PGA) which indicates that G-PGA successfully maximizes the adversarial features while minimizes the true class features. We quantify this by measuring the $L_1$ distance between attention maps of the clean and adversarial images generated with and without guidance. Our G-PGA creates more attention dispersion (Fig. \ref{fig:attn_maps_quantity}).

\subsection{Optimal Surrogate}
\begin{figure}[!t]
\centering
  \begin{minipage}{.48\columnwidth}
  	\centering
    \includegraphics[ width=\linewidth,keepaspectratio]{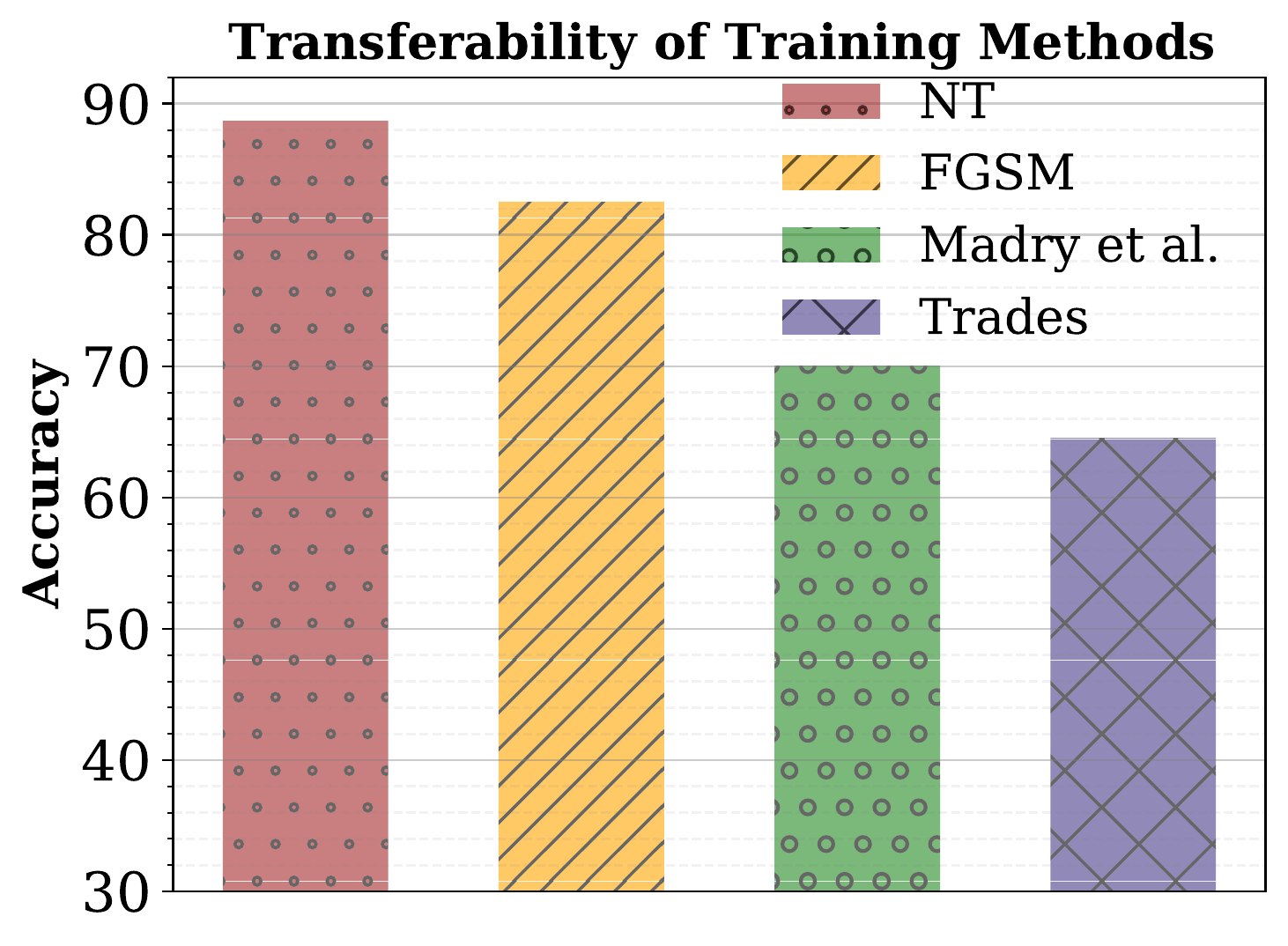}
  \end{minipage}
  \begin{minipage}{.48\columnwidth}
  	\centering
    \includegraphics[width=\linewidth, height=3.5cm]{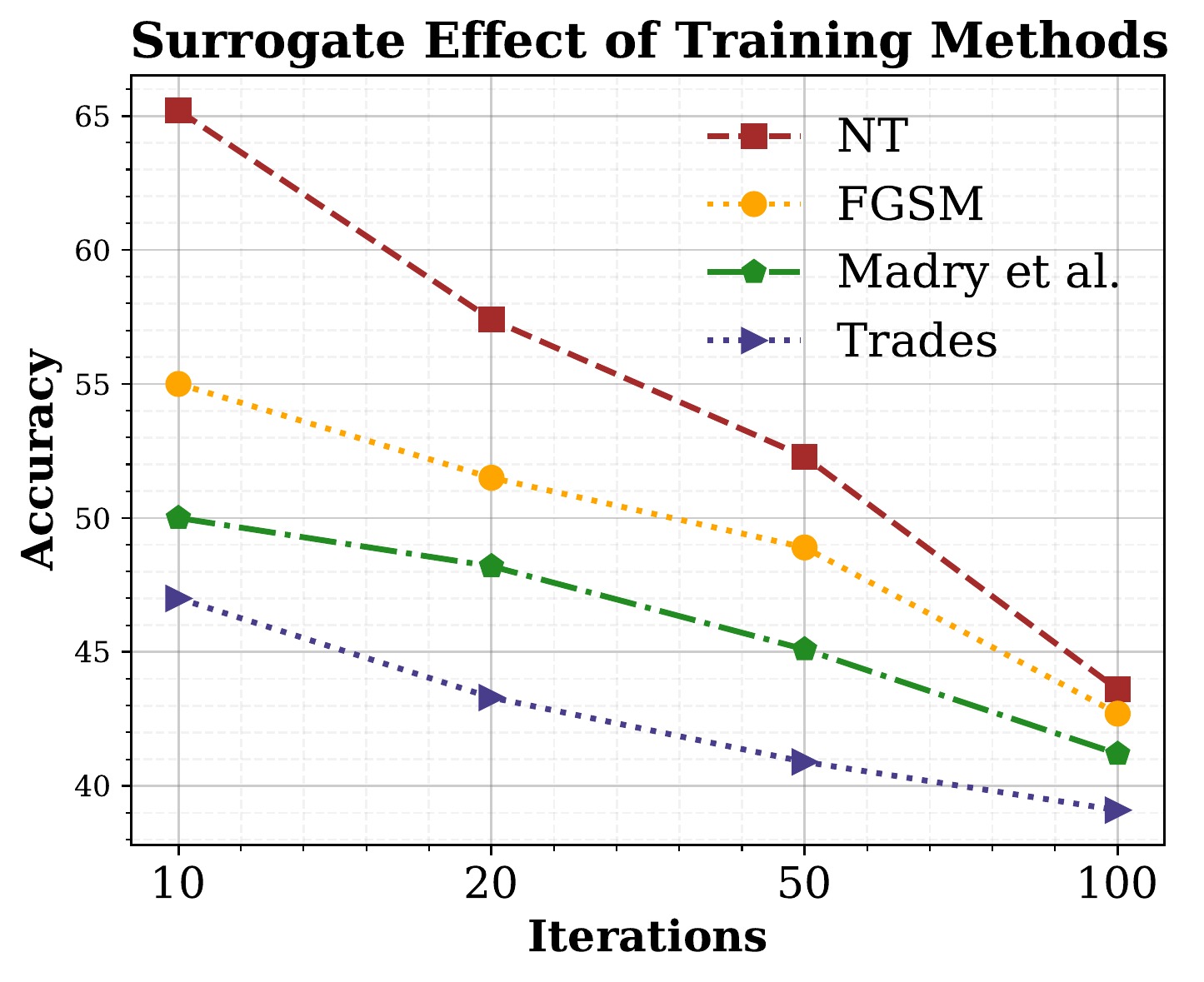}
  \end{minipage}
  \caption{Robustness of (FS \cite{feature_scatter}) in terms of Top-1 accuracy (\%) on CIFAR10 dataset. WideResNet is used in the experiment. NT represents natural training. Trades is trained at $\beta=1$. \textbf{Left} plot shows \emph{{black-box}} robustness of feature scattering against adversarial attack (PGD, 10 iterations, step size2/255) transferred from models trained using different training methods.
  \textbf{Right} plot shows \emph{{white-box}} robustness of FS when the different training mechanism used as a surrogate in our proposed attack (Algo. \ref{alg:guided_attack}, 10 iterations).}
  \label{fig:surrogate_effect_of_trainings}
\end{figure}

We use Trades \cite{Zhang2019theoretically} framework to find the effective surrogate model. Trades introduces a tradeoff parameter, $\beta$. When $\beta=0$, Trades converges to natural training. When $\beta$ is increased, the model becomes adversarially stronger at the expense of clean accuracy. We train ResNet18 using \cite{Zhang2019theoretically} at $\beta \in \{0,1,6\}$ and use them as surrogate models to observe the surrogate effect on another ResNet18 trained using Mask-AT (Algo. \ref{alg:masking_at} with $\eta=6, \delta=0.75$). G-PGA successfully provides the required guidance to overcome gradient masking using surrogate information from naturally ($\beta=0$) as well as adversarially trained models (Fig. \ref{fig:analysis_match_and_deceive}). However, for fixed iterations, surrogate information at $\beta=1$ is the most effective. We further analyze guidance provided by different training methods. Fig.~\ref{fig:surrogate_effect_of_trainings} shows that models trained using Trades \cite{Zhang2019theoretically} have better black-box transferability and provide faster convergence when used in our attack. However, our approach can successfully exploit surrogate information from naturally trained models as well but requires more attack iterations (Fig. \ref{fig:surrogate_effect_of_trainings}). In light of this experiment (Fig. \ref{fig:analysis_match_and_deceive}), we use the surrogate model trained using \cite{Zhang2019theoretically} at $\beta=1$. Our analysis (Fig. \ref{fig:analysis_match_and_deceive}) shows that the surrogate model trained via Trades ($\beta=1$) is the optimal condition for better guidance but not the necessary condition for an effective guided attack (Table \ref{tab: imagenet_results}).

\subsubsection{Different Surrogate Models}
Our guided mechanism is not constrained by the same architecture of original and surrogate model that is G-PGA can achieve guidance from a totally different architecture than the original (under-attacked) model. We present this analysis in Table \ref{tab: different_surrogates}.

\begin{table}[!htb]
	\centering
		\begin{tabular}{ c c c c c c c c c |c }
		\toprule
			\rowcolor{Gray}
			Original & Surrogate & G-PGA ($\downarrow$) \\
			 \midrule
			WideResNet  & ResNet18 & \textbf{36.0}\\
			ResNet18 & NiN & \textbf{30.1} \\
			 ResNet18  & WideResNet & \textbf{28.3}\\
			\bottomrule
	\end{tabular}
	\caption{ Effect of different surrogate models on G-PGA. Large capacity models such WideResNet helps more as surrogate but also difficult to attack. Evaluations (\% Top-1) are presented against FS \cite{feature_scatter} defense on CIFAR-10 test set. NiN represents "Network In Networks" model.}
	\label{tab: different_surrogates}
\end{table}

\subsubsection{Different Contrastive Directional Losses}
\begin{figure}[!t]
\centering
  \begin{minipage}{.48\linewidth}
  	\centering
    \includegraphics[ width=\linewidth,keepaspectratio]{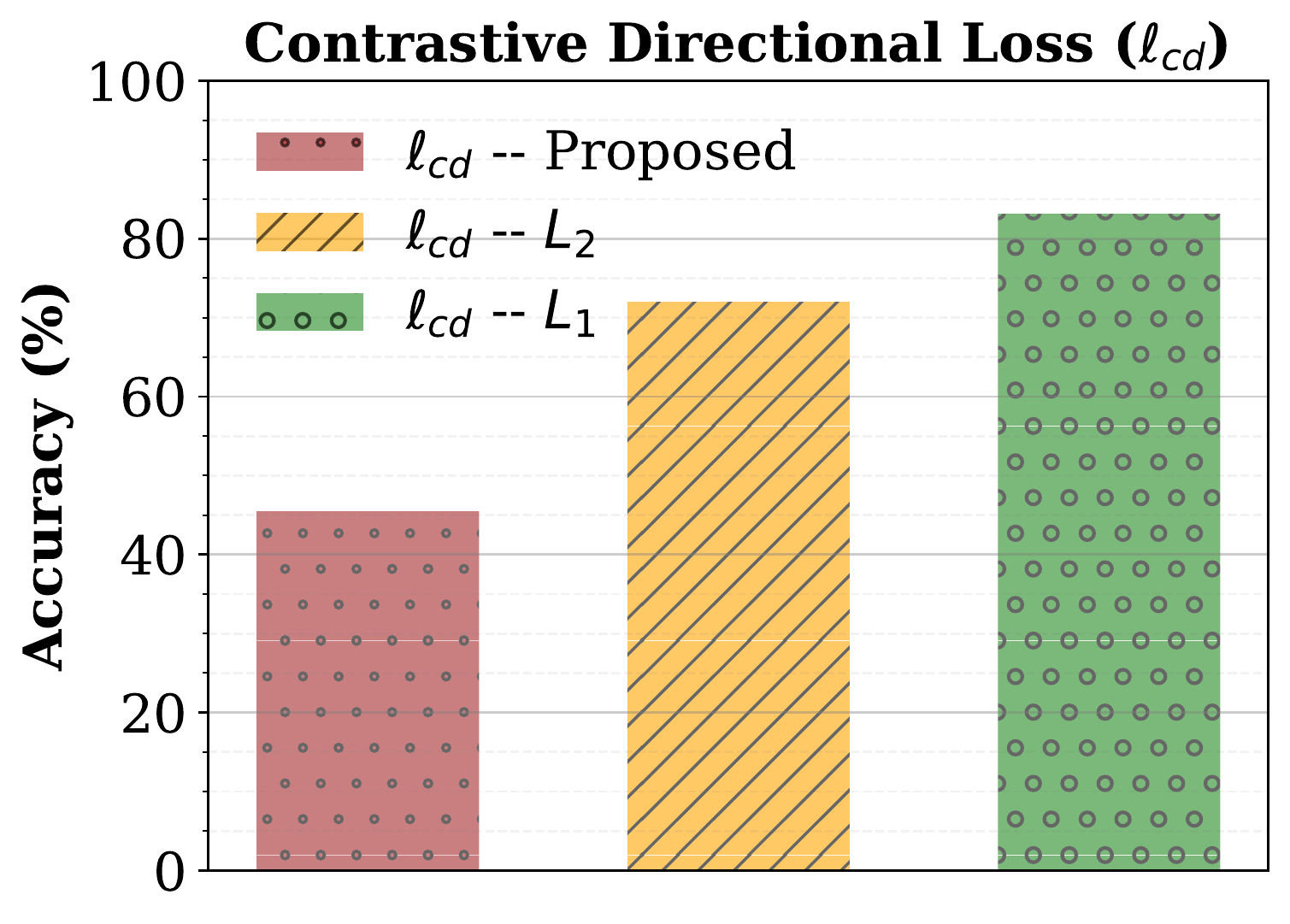}
    \scriptsize  (a) \textbf{Effect of model $h$ ($\beta=0$)}
  \end{minipage}
  \begin{minipage}{.48\linewidth}
  	\centering
    \includegraphics[width=\linewidth, keepaspectratio]{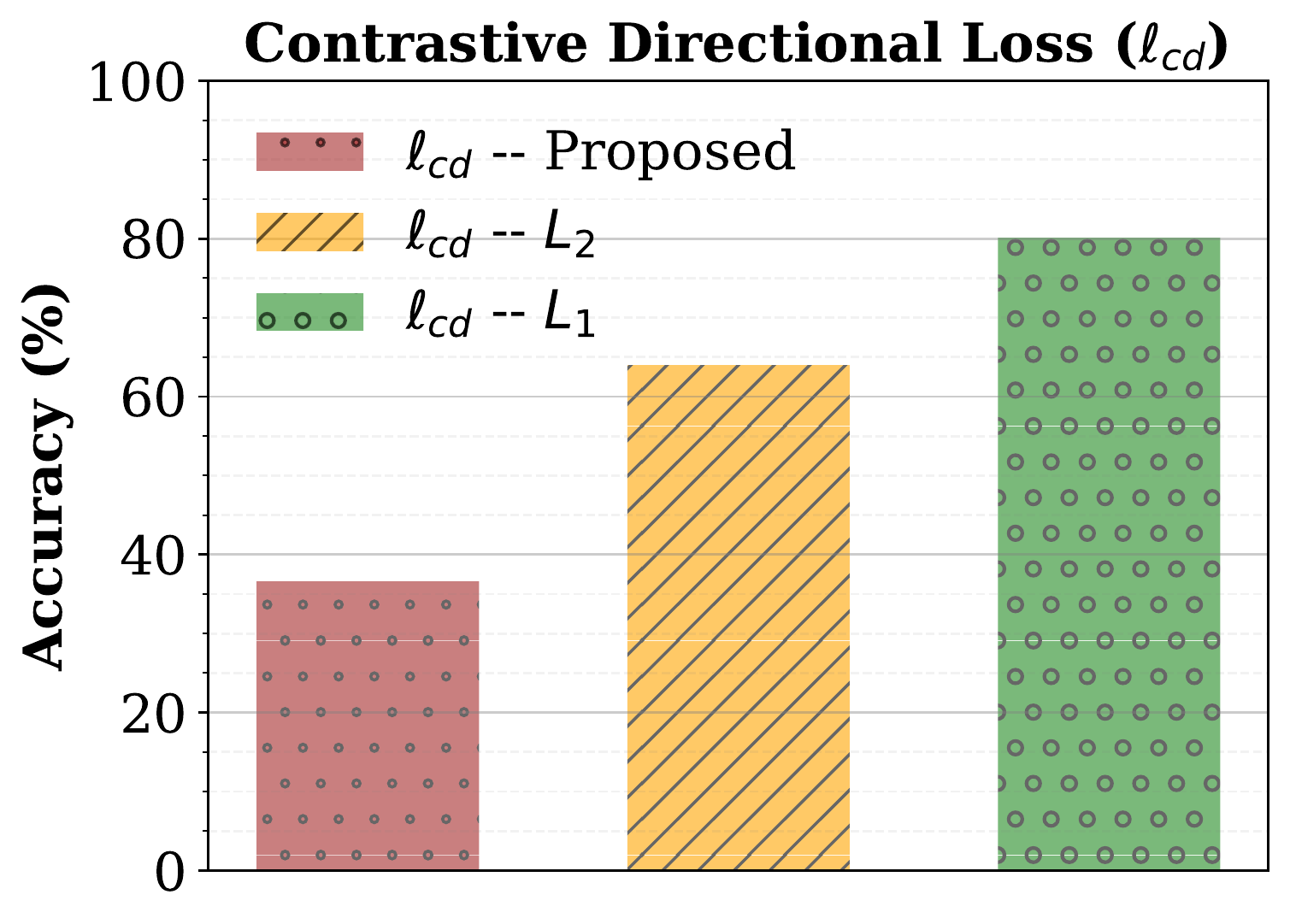}
    \scriptsize  (b) \textbf{Effect of model $h$ ($\beta=1$)}
  \end{minipage}
  \caption{We evaluate the performance of different metrics within our proposed contrastive loss ($\ell_{cd}$, Eq. \ref{eq:contrastive_directional_loss} ). Our proposed objective performs significantly better than $L_1$ or $L_2$ measures with $\ell_{cd}$ (\emph{lower is better}). Results are on CIFAR10 test set. }
\label{fig:lcd_ce_l2_l1}
\end{figure}

When the original model hides gradients or provides noisy gradient estimation then contrastive directional loss pushes the optimizer to  move  along  the  adversarial direction defined by the masking free surrogate model. This is demonstrated by our results as well (Tables \ref{tab:main-results} and \ref{tab:comparison_with_AA}, and Fig. \ref{fig:defenses_vs_GPGA}). We compare our proposed formulation (Eq. \ref{eq:contrastive_directional_loss}) with $L_1$ and $L_2$ losses in contrastive directional loss (Fig. \ref{fig:lcd_ce_l2_l1}). Our proposed objective provides favorable results in contrastive directional loss.

\section{Conclusion}
Gradient masking is a recurring phenomenon in evaluation of adversarial robustness. Our work sheds light on the elusive robustness caused by the label smoothing. We design an adversarial training algorithm that artificially increases model robustness by hiding gradients with the help of label smoothing. We then propose a new attack (G-PGA) based on the concept of guided optimization that exposes gradient masking within a few attack iterations.  Our attack approach is based on a novel redirection and rescaling mechanism that uses guidance from a surrogate teacher model on a given target model. G-PGA finds useful adversarial directions that ultimately help to skip local minima during attack optimization. The redirection is achieved with a contrastive directional loss, while rescaling is performed using a normalized CE objective. We hope our findings can act as a guide on the future use of label smoothing in adversarial training and devising diagnostic tools to catch masking.

\appendices
\section{Proof of gradients}
Here we provide proof for the gradients formulae of $\ell_{cd}$:
\begin{align}
    \frac{\partial \ell_{cd}}{\partial \bm{u}} =  - \frac{ (\bm{z}'- \bm{v}') \exp(\bm{v}'\cdot \bm{u}') (\bm{I} - \bm{u}' \cdot \bm{u}'^{\top})}{ \| \bm{u}\| (\exp(\bm{v}'\cdot \bm{u}') + \exp(\bm{u}' \cdot \bm{z}') )  }, 
    \\
    \frac{\partial \ell_{cd}}{\partial \bm{z}} = - \frac{ \bm{u}'   \exp(\bm{v}'\cdot \bm{u}') (\bm{I} - \bm{z}' \cdot \bm{z}'^{\top}) }{ \| \bm{z}\|(\exp(\bm{v}'\cdot \bm{u}') + \exp(\bm{u}' \cdot \bm{z}') ) }.
\end{align}

\noindent\textbf{\emph{Proof:}}
\begin{align}
   \frac{\partial \ell_{cd}}{\partial \bm{u}} = \frac{\partial \bm{u}' }{\partial \bm{u}} \cdot \frac{\partial \ell_{cd} }{ \partial \bm{u}'},
\end{align}
\begin{align}
    \frac{\partial \ell_{cd} }{ \partial \bm{u}'} & = \frac{\partial }{\partial  \bm{u}'}  \left[- \log \frac{ \exp(\bm{u}' \cdot \bm{z}')}{\exp(\bm{v}'\cdot \bm{u}') + \exp(\bm{u}' \cdot \bm{z}') } \right] \notag\\
    & = - \left( \frac{ \exp(\bm{v}'\cdot \bm{u}') + \exp(\bm{u}' \cdot \bm{z}') }{ \exp(\bm{u}' \cdot \bm{z}')}\right)  \notag\\
    & \frac{\partial }{\partial  \bm{u}'}  \left(  \frac{ \exp(\bm{u}' \cdot \bm{z}')}{\exp(\bm{v}'\cdot \bm{u}') + \exp(\bm{u}' \cdot \bm{z}') } \right), \notag\\
    & = -  \frac{ 1}{\exp(\bm{u}' \cdot \bm{z}')(\exp(\bm{v}'\cdot \bm{u}') + \exp(\bm{u}' \cdot \bm{z}')) } ( \notag\\ & \exp(\bm{v}'\cdot \bm{u}') + \exp(\bm{u}' \cdot \bm{z}') ) \frac{\partial}{\partial \bm{u}'} \exp(\bm{u}' \cdot \bm{z}') - \notag\\
    & \exp(\bm{u}' \cdot \bm{z}')\frac{\partial }{\partial \bm{u}'}(\exp(\bm{v}'\cdot \bm{u}') + \exp(\bm{u}' \cdot \bm{z}') )  \notag\\
    & = - \frac{1}{\exp(\bm{u}' \cdot \bm{z}')(\exp(\bm{v}'\cdot \bm{u}') + \exp(\bm{u}' \cdot \bm{z}')) } ( \notag\\
    &\bm{z}' \exp(\bm{u}' \cdot \bm{z}')(\exp(\bm{v}'\cdot \bm{u}') + \exp(\bm{u}' \cdot \bm{z}')) - \notag\\
    & \bm{z}' \exp(2 \cdot \bm{u}' \cdot \bm{z}') - \bm{v}'\exp(\bm{u}' \cdot \bm{z}')\exp(\bm{v}'\cdot \bm{u}') ) \notag\\
    & = - \frac{ (\bm{z}' - \bm{v}') \exp(\bm{v}'\cdot \bm{u}') }{ \exp(\bm{v}'\cdot \bm{u}') + \exp(\bm{u}' \cdot \bm{z}') }.
\end{align}

\begin{align}
    \frac{\partial \bm{u}'}{ \partial \bm{u} } 
    & =   \frac{\partial }{ \partial \bm{u} } \left( \frac{\bm{u}}{ \| \bm{u} \| } \right) \notag\\
    & = \frac{ \| \bm{u} \| - \bm{u} \frac{\partial }{ \partial \bm{u}} \| \bm{u}\| }{ \| \bm{u}\|^2 } =  \frac{ 1 }{  \| \bm{u}\| } -  \frac{ \bm{u} \frac{\partial }{ \partial \bm{u}} \sqrt{ \bm{u} \cdot \bm{u}^{\top}} }{ \| \bm{u}\|^2 } \notag\\
    & = \frac{1}{\| \bm{u}\|}(\bm{I} - \bm{u}' \cdot \bm{u}'^{\top}).
\end{align}
Then,
\begin{align}
    \frac{\partial \ell_{cd}}{\partial \bm{u}} =  - \frac{ (\bm{z}'- \bm{v}') \exp(\bm{v}'\cdot \bm{u}') (\bm{I} - \bm{u}' \cdot \bm{u}'^{\top})}{ \| \bm{u}\| (\exp(\bm{v}'\cdot \bm{u}') + \exp(\bm{u}' \cdot \bm{z}') )  }.
\end{align}
Since, $\bm{z} = h(\tilde{\bm{x}})$ also depends on the adversarial input, we can calculate $\frac{\partial \ell_{cd}}{\partial \bm{z}}$ as follows,
\begin{align}
    \frac{\partial \ell_{cd}}{\partial \bm{z}} = \frac{\partial \bm{z}'} {\partial \bm{z}} \cdot \frac{\partial \ell_{cd}}{\partial \bm{z}'},
\end{align}
\begin{align}
    & \frac{\partial \ell_{cd}}{\partial \bm{z}'}
     = \frac{\partial }{\partial  \bm{z}'}  \left[- \log \frac{ \exp(\bm{u}' \cdot \bm{z}')}{\exp(\bm{v}'\cdot \bm{u}') + \exp(\bm{u}' \cdot \bm{z}') } \right] \notag\\
    & = - \left( \frac{ \exp(\bm{v}'\cdot \bm{u}') + \exp(\bm{u}' \cdot \bm{z}') }{ \exp(\bm{u}' \cdot \bm{z}')}\right)  \frac{\partial }{\partial  \bm{z}'}  \notag \\
    & \left(  \frac{ \exp(\bm{u}' \cdot \bm{z}')}{\exp(\bm{v}'\cdot \bm{u}') + \exp(\bm{u}' \cdot \bm{z}') } \right), \notag\\
    & = -  \frac{ 1}{\exp(\bm{u}' \cdot \bm{z}')(\exp(\bm{v}'\cdot \bm{u}') + \exp(\bm{u}' \cdot \bm{z}')) } (\exp(\bm{v}'\cdot \bm{u}') + \notag \\
    & \exp(\bm{u}' \cdot \bm{z}') ) \frac{\partial}{\partial \bm{z}'} \exp(\bm{u}' \cdot \bm{z}')- \exp(\bm{u}' \cdot \bm{z}')\frac{\partial }{\partial \bm{z}'}(\exp(\bm{v}'\cdot \bm{u}') \notag\\
    & + \exp(\bm{u}' \cdot \bm{z}') )  \notag\\
    & = - \frac{ \bm{u}' (\exp(\bm{v}'\cdot \bm{u}') + \exp(\bm{u}' \cdot \bm{z}')) -   \bm{u}' \exp( \bm{u}' \cdot \bm{z}')  }{\exp(\bm{v}'\cdot \bm{u}') + \exp(\bm{u}' \cdot \bm{z}') } \notag\\
    & = - \frac{ \bm{u}' \exp(\bm{v}'\cdot \bm{u}') }{ \exp(\bm{v}'\cdot \bm{u}') + \exp(\bm{u}' \cdot \bm{z}') }.
\end{align}
which completes the proof. 

\bibliographystyle{IEEEtran}
\bibliography{mybibfile}

%









\begin{IEEEbiography}[{\includegraphics[width=1in,height=1.5in,clip,keepaspectratio]{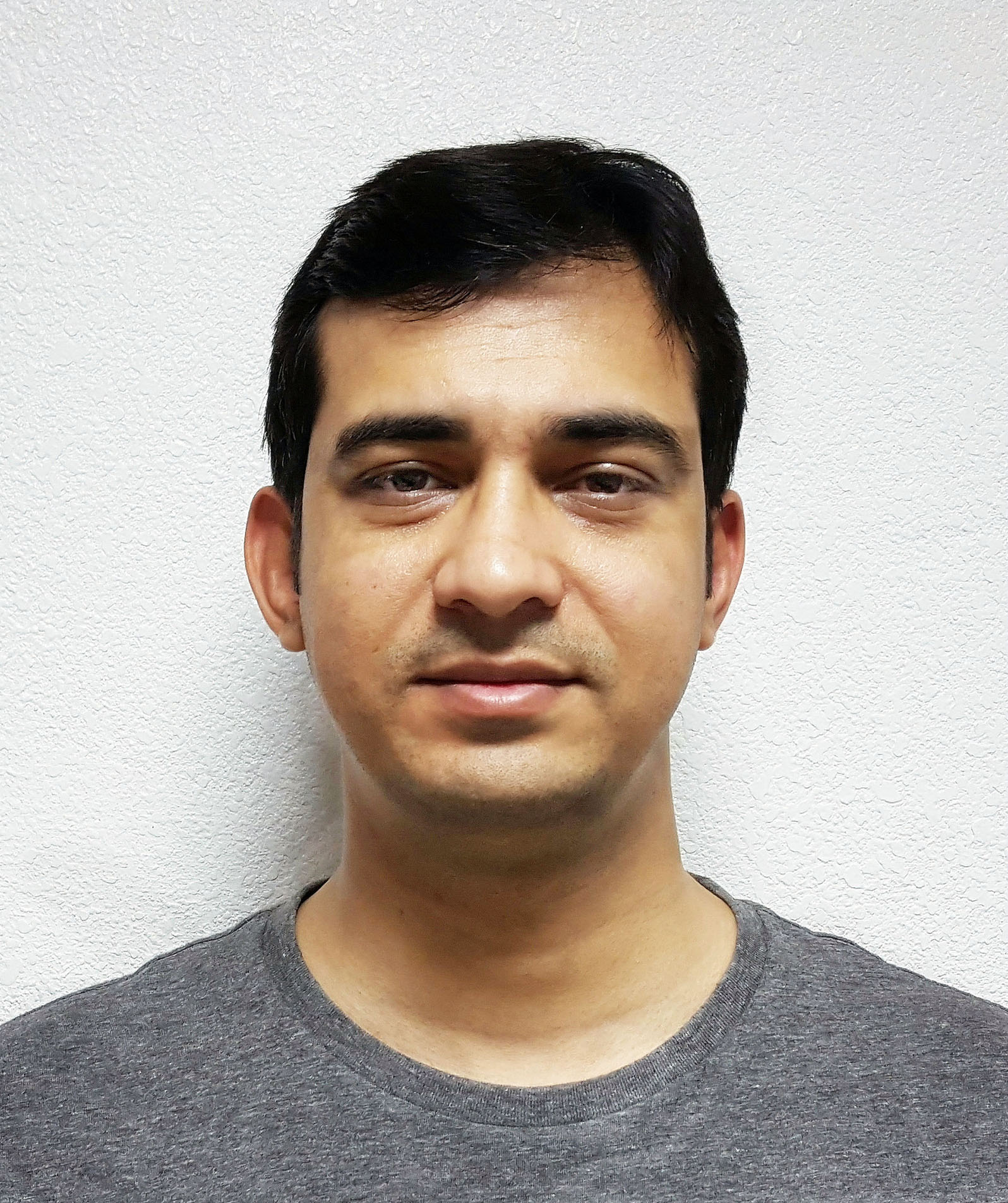}}]
{Muzammal Naseer}
received the Ph.D. degree from the Australian National University (ANU) in 2022, where he was the recipient of a competitive postgraduate scholarship. He is currently a postdoctoral researcher at Mohamed Bin Zayed University of Artificial Intelligence. He served as a researcher at Data61, CSIRO, and Inception Institute of Artificial Intelligence from 2018-2020. He has published at well recognized machine learning and computer vision venues including NeurIPS, ICLR, ICCV, and CVPR with two Oral and two spotlight presentations. He received student travel award from NeurIPS in 2019. He received Gold Medal for outstanding performance in the B.Sc. degree.
\end{IEEEbiography}\vspace{-2em}

\begin{IEEEbiography}[{\includegraphics[width=1in,height=1.5in,clip,keepaspectratio]{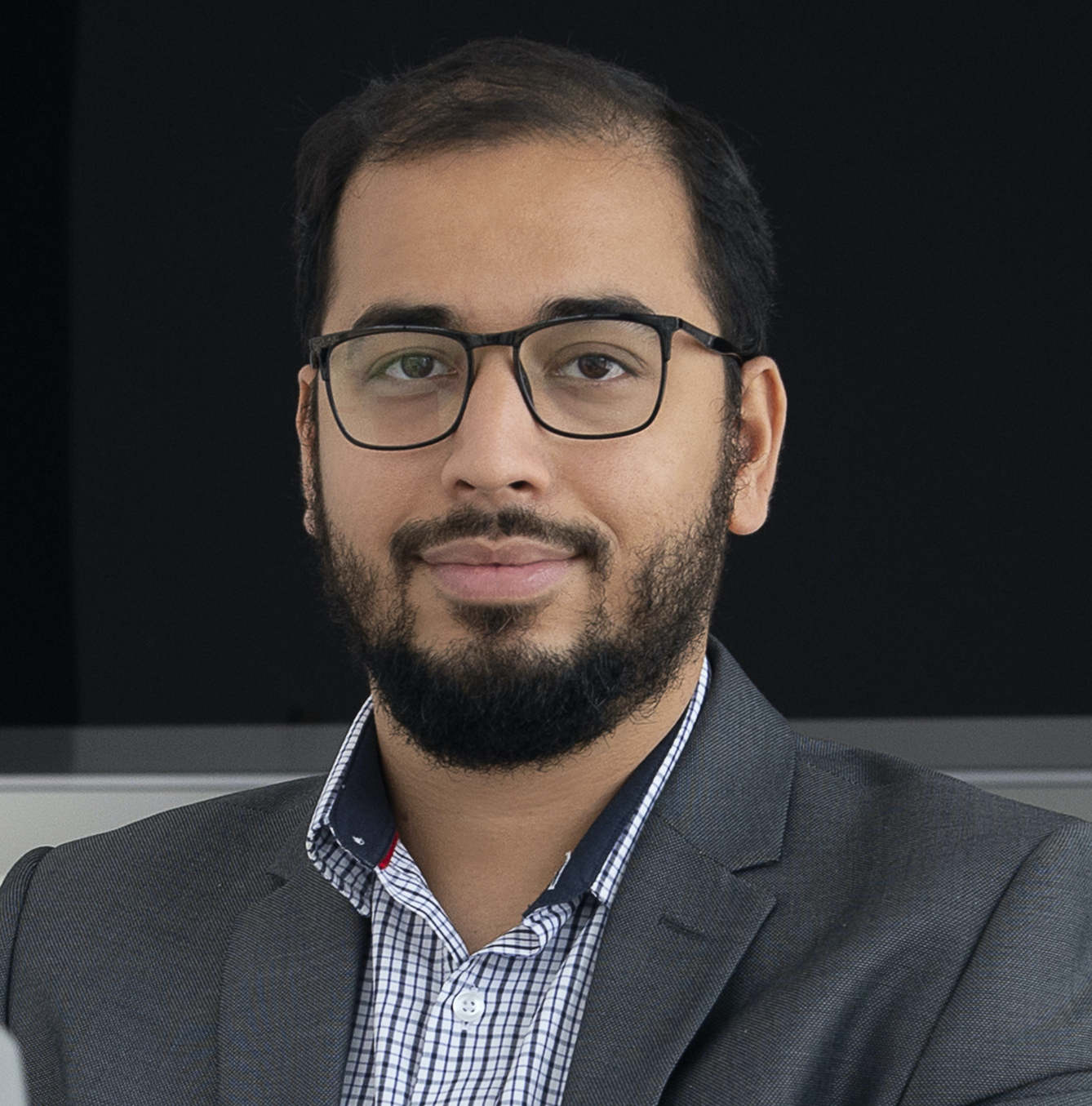}}]
{Salman Khan} (M'14-SM'22)
 received the Ph.D. degree from The University of Western Australia, in 2016. His Ph.D. thesis received an honorable mention on the Deans List Award. From 2016 to 2018, he was a Research Scientist with Data61, CSIRO. He was a Senior Scientist with Inception Institute of Artificial Intelligence from 2018-2020. He is currently acting as an Associate Professor at Mohamed Bin Zayed University of Artificial Intelligence, since 2020, and an Adjunct Lecturer with Australian National University, since 2016. He has served as a program committee member for several premier conferences, including CVPR, ICCV, IROS, ICRA, ICLR, NeurIPS and ECCV. In 2019, he was awarded the outstanding reviewer award at CVPR and the best paper award at ICPRAM 2020. His research interests include computer vision and machine learning.
 \end{IEEEbiography}

\begin{IEEEbiography}[{\includegraphics[width=1in,height=1.5in,clip,keepaspectratio]{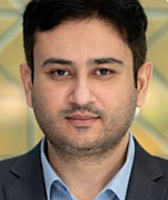}}]
{Fahad Shahbaz Khan}
is a faculty member at MBZUAI, United Arab Emirates and Linköping University, Sweden. He received the M.Sc. degree in Intelligent Systems Design from Chalmers University of Technology, Sweden and a Ph.D. degree in Computer Vision from Autonomous University of Barcelona, Spain. He has achieved top ranks on various international challenges (Visual Object Tracking VOT: 1st 2014 and 2018, 2nd 2015, 1st 2016; VOT-TIR: 1st 2015 and 2016; OpenCV Tracking: 1st 2015; 1st PASCAL VOC 2010). He received the best paper award in the computer vision track at IEEE ICPR 2016. His research interests include a wide range of topics within computer vision and machine learning, such as object recognition, object detection, action recognition and visual tracking. He has published over 100 conference papers, journal articles, and book contributions in these areas. He has served as a guest editor of IEEE Transactions on Pattern Analysis and Machine Intelligence, IEEE Transactions on Neural Networks and Learning Systems and is an associate editor of Image and Vision Computing Journal. He serves as a regular program committee member for leading computer vision conferences such as CVPR, ICCV, and ECCV.
\end{IEEEbiography}

\begin{IEEEbiography}[{\includegraphics[width=1in,height=1.5in,clip,keepaspectratio]{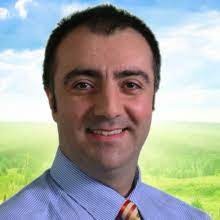}}]
{Fatih Porikli}
(M’96–SM’04–F’14) received the Ph.D. degree from New York University, in 2002. He was a Distinguished Research Scientist with Mitsubishi Electric Research Laboratories. He was a full tenured Professor in the Research School of Engineering, Australian National University, and a Chief Scientist with the Global Media Technologies Lab, Huawei, Santa Clara. He is currently the Global Lead of Perception at Qualcomm. He has authored over 300 publications, co-edited two books, and invented 66 patents. His research interests include computer vision, pattern recognition, manifold learning, image enhancement, robust and sparse optimization, and online learning with commercial applications in video surveillance, car navigation, robotics, satellite, and medical systems. He was a recipient of the Research and Development 100 Scientist of the Year Award, in 2006. He received five best paper awards at premier IEEE conferences and five other professional prizes. He is serving as an associate editor for several journals for the past 12 years. He has also served in the organizing committees of several flagship conferences, including ICCV, ECCV, and CVPR.
\end{IEEEbiography}
\vfill

\end{document}

%% file: plots/analysis_on_gradient_masking.tex
\begin{figure*}[!t]
\centering
  \begin{minipage}{.230\textwidth}
  	\centering
    \includegraphics[ width=\linewidth,keepaspectratio, clip=true,trim=0 0 0 0mm]{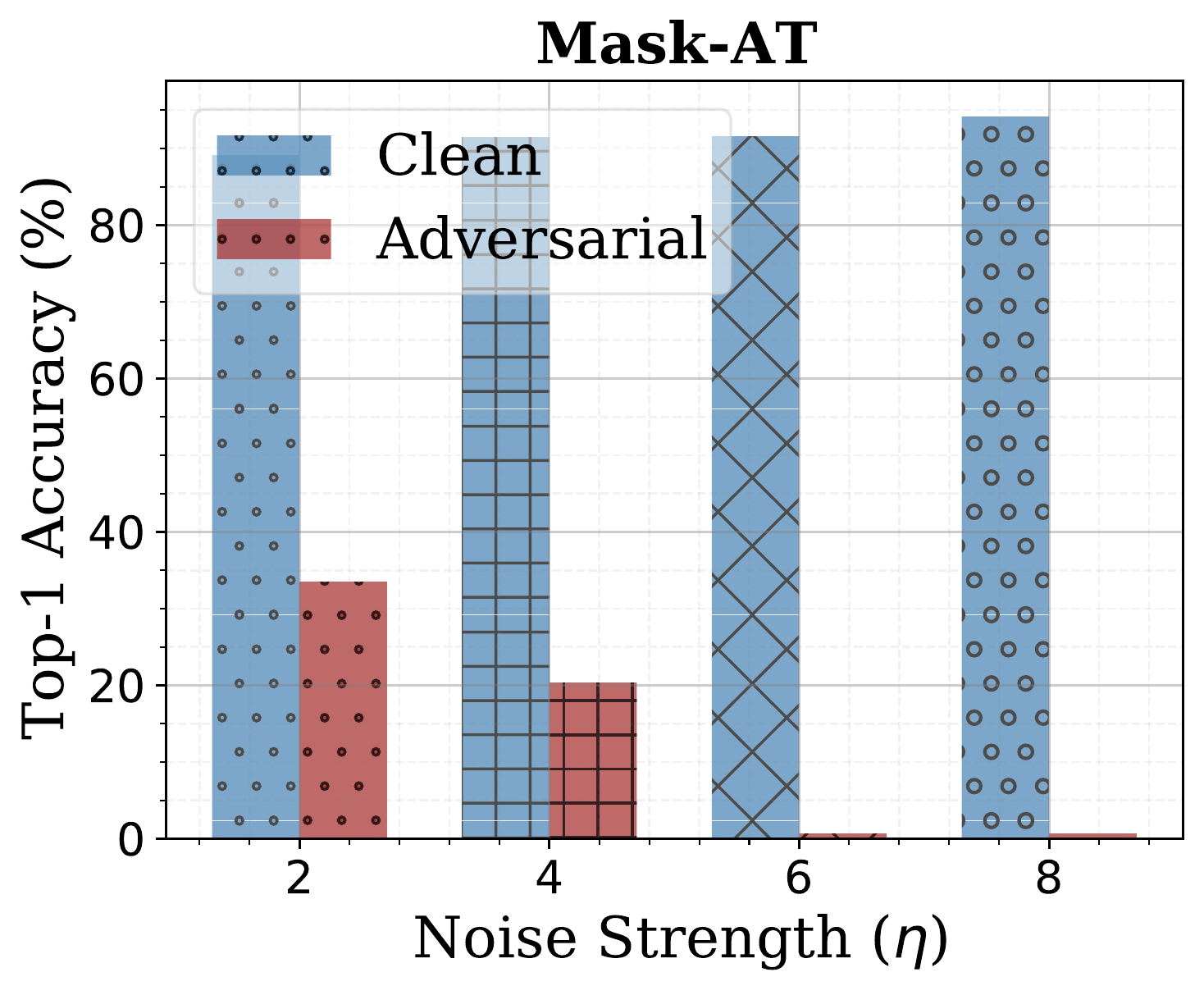}
    \tiny (a) \textbf{Effect of $\eta$ with $\delta{=}0$}
  \end{minipage}
  \begin{minipage}{.230\textwidth}
  	\centering
    \includegraphics[width=\linewidth, keepaspectratio, clip=true,trim=0 0 0 0mm]{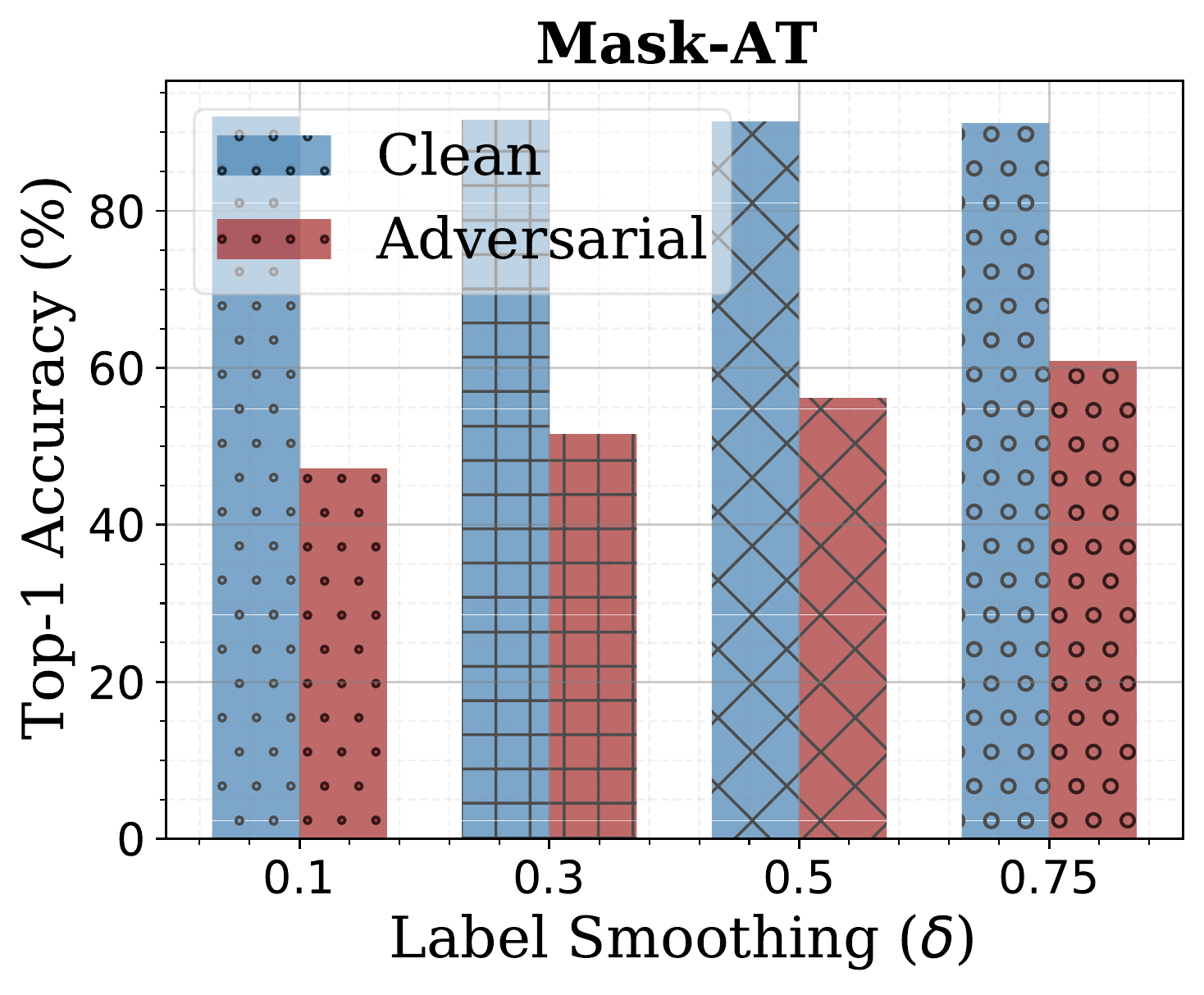}
    \tiny (b) \textbf{Effect of $\delta$ with $\eta{=}4$}
  \end{minipage}
  \begin{minipage}{.230\textwidth}
  	\centering
    \includegraphics[width=\linewidth, keepaspectratio, clip=true,trim=0 0 0 0mm]{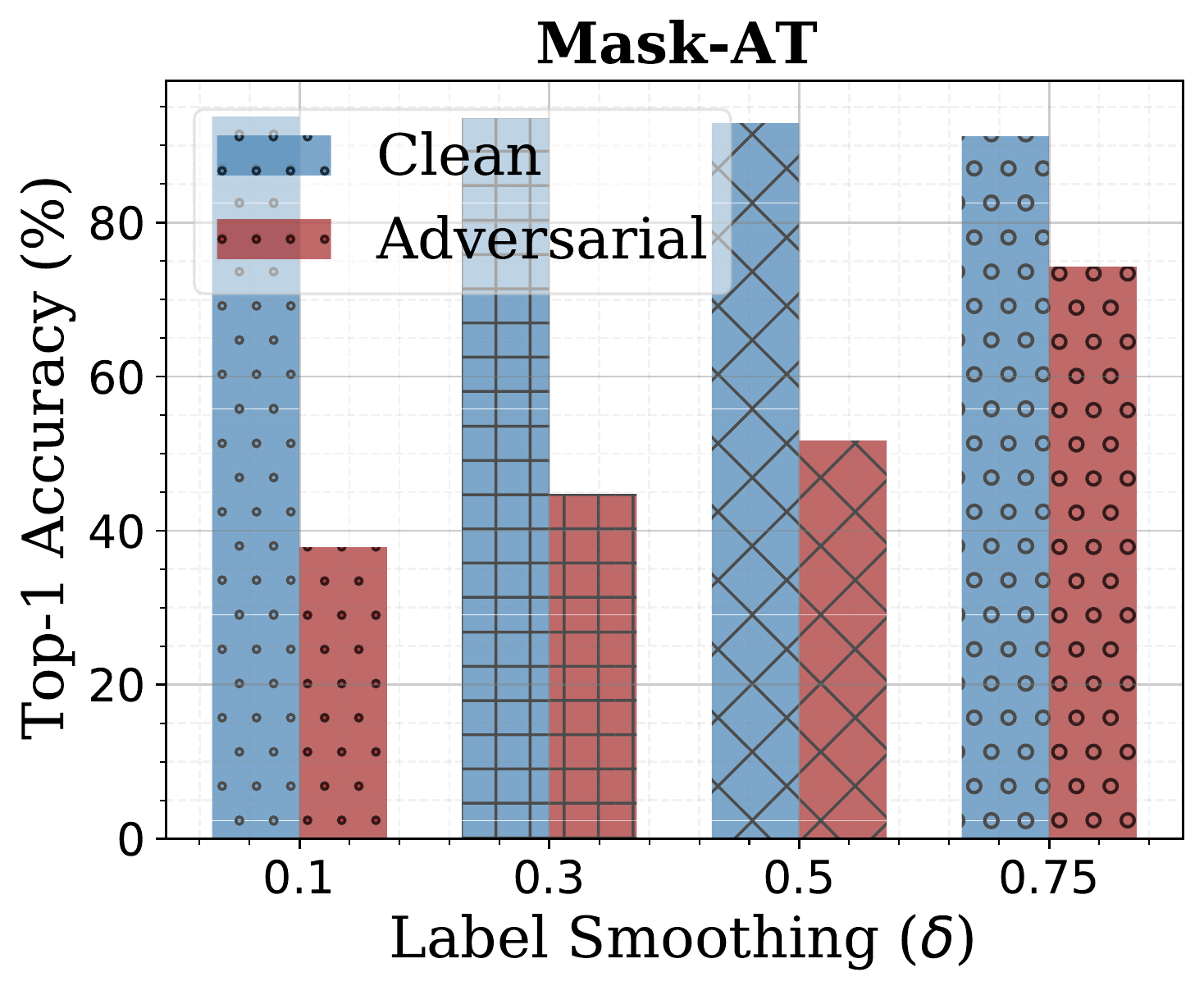}\
    \tiny (c) \textbf{Effect of $\delta$ with $\eta{=}6$}
  \end{minipage}
  \begin{minipage}{.230\textwidth}
  	\centering
    \includegraphics[width=\linewidth, keepaspectratio, clip=true,trim=0 0 0 0mm]{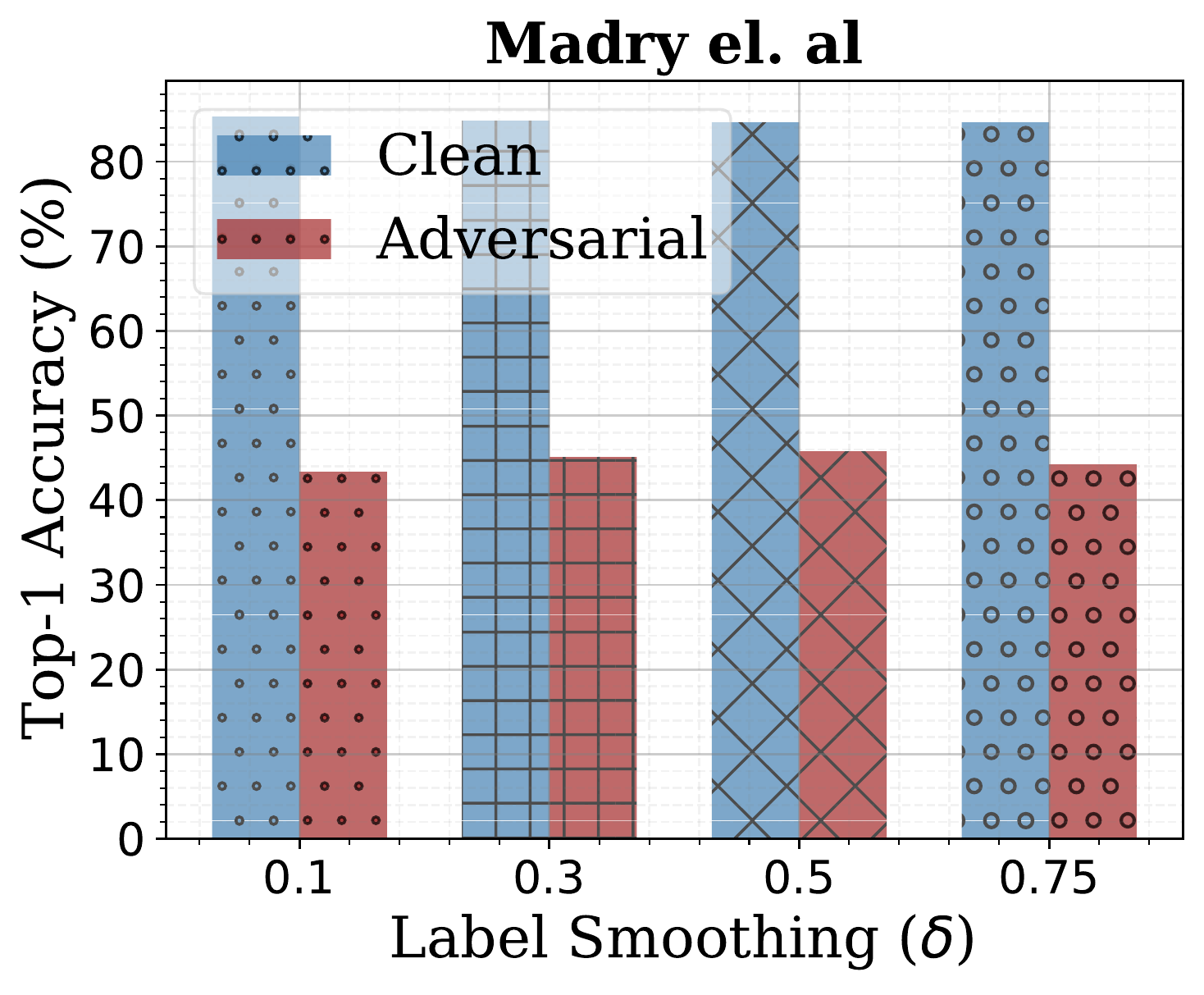}\
    \tiny (d) \textbf{Effect of $\delta$ with $\eta{=}4$}
  \end{minipage}
\caption{\emph{Analysis on Gradient Masking:} Clean vs. Adversarial accuracy  (\% Top-1) of ResNet18 on CIFAR10 is reported against PGD attack with 20 iterations. PGD successfully highlights true robustness in the absence of gradient masking \eg, when models are trained on FGSM (a) or using Madry's method \cite{madry2018towards} (d). However, PGD erroneously indicates very high robustness of models trained on FGSM adversaries with label smoothing (b-c). Thus, with the help of label smoothing only, adversarial robustness of Mask-AT (Algo. \ref{alg:masking_at}) increases from zero to around 75\% while maintaining the clean accuracy (plot (c): $\eta=6$, $\delta=0.75$). This signifies how the use of a certain component during adversarial training (such as label smoothing in this case) can paralyze an attack and shows fake robustness. A solution to this problem is to guide the attack optimization (Sec.~\ref{sec:methodology}).}
\label{fig:analysis_mask_at}
\end{figure*}

%% file: tables/main_table.tex
\begin{table*}[!t]
		\centering\small
			\setlength{\tabcolsep}{9pt}
				\scalebox{0.95}[0.95]{
				\begin{tabular*}{\textwidth}{l|c|c|ccc|ccc|ccc}
					\toprule
					\rowcolor{Gray}
				\multicolumn{12}{c}{\small\textbf{CIFAR10.} Perturbation budget is 8/255 in $\ell_{\infty}$ norm.}\\
				\midrule[0.4mm]
				\multirow{1}{*}{Defense} &\multirow{1}{*}{$\delta$}  & \multirow{1}{*}{Clean} & \multicolumn{3}{c|}{PGD}&\multicolumn{3}{c|}{CW}&\multicolumn{3}{c}{Proposed Attack}\\
				\cline{4-12} 
					&&&   10 & 20 & 100 & 10 & 20 & 100&10&20&50\\
					\midrule
					Madry \etal\cite{madry2018towards} (\texttt{ICLR'18}) & 0.5&86.2&49.3&47.5&46.8&47.8&46.4&45.8&46.6&45.2&\textbf{44.7} \\
					Trades \cite{Zhang2019theoretically} (\texttt{ICML'19}) & 0.5&86.0&54.3&53.4&53.1&52.6&52.1&52.0&53.4&52.5&\textbf{51.9}\\
					FS \cite{feature_scatter} (\texttt{NeuRIPS'19})& 0.5&90.0&70.9&70.5&68.6&62.6&62.4&60.6&47.0&43.5&\textbf{40.9}\\
					AvMix \cite{lee2020adversarial} (\texttt{CVPR'20}) & 0.5-0.7&93.2&73.9&72.1&70.4&65.4&62.5&59.7&57.5&51.5&\textbf{46.4} \\
					Mask-AT  (Ours) & 0.75 &93.4&75.1&74.5&72.9&71.4&68.4&64.6&41.2&34.7&\textbf{29.7}\\
					\hline \midrule[0.4mm]
		            	\rowcolor{Gray}
			       \multicolumn{12}{c}{\small\textbf{CIFAR100.} Perturbation budget is 8/255 in $\ell_{\infty}$ norm.}\\
			       \midrule[0.4mm]
					Madry \etal\cite{madry2018towards} (\texttt{ICLR'18}) & 0.5&60.2&26.6&25.9&25.6&24.1&23.7&23.6&25.4&24.9&\textbf{23.4} \\
					Trades \cite{Zhang2019theoretically} (\texttt{ICML'19}) & 0.5&59.1&28.7&28.4&28.3&26.1&25.8&25.7&27.4&26.1&\textbf{25.4}\\
					FS \cite{feature_scatter} (\texttt{NeuRIPS'19})& 0.5&74.7&43.0&42.9&42.0&26.5&25.8&24.6&2.2&1.5&\textbf{1.1}\\
					AvMix \cite{lee2020adversarial} (\texttt{CVPR'20}) & 0.5-0.7&74.5&45.4&44.8&43.2&31.2&29.7&27.4&4.4&3.0&\textbf{1.9} \\
					Mask-AT (Ours) & 0.75&74.2&42.2&41.2&39.5&28.6&27.1&25.2&2.9&1.9&\textbf{1.6}\\
					\hline \midrule[0.4mm]
		            \rowcolor{Gray}
	                \multicolumn{12}{c}{\small\textbf{SVHN.} Perturbation budget is 8/255 in $\ell_{\infty}$ norm.}\\
			       \midrule[0.4mm]
					Madry \etal\cite{madry2018towards} (\texttt{ICLR'18}) & 0.5 & 96.1 & 52.1& 51.5&51.3&49.2&48.6&47.0&49.0&47.4&\textbf{46.3}\\
					Trades \cite{Zhang2019theoretically} (\texttt{ICML'19}) & 0.5&95.0&62.0&61.8&61.5&58.9&56.7&56.1&58.2&57.3&\textbf{55.9}\\
					FS \cite{feature_scatter} (\texttt{NeuRIPS'19})& 0.5&96.0&60.6&46.1&25.7&53.4&37.7&19.3&48.4&34.5&\textbf{19.0}\\
					AvMix \cite{lee2020adversarial} (\texttt{CVPR'20}) & 0.5-0.7&96.7&77.2&66.0&40.5&73.2&61.5&37.3&64.5&53.7&\textbf{36.0} \\
					Mask-AT  (Ours) & 0.75&96.5&74.2&66.1&52.6&70.3&60.4&46.6&54.1&44.6&\textbf{34.4}\\
					\bottomrule
			\end{tabular*}}
		\caption{\centering Evaluation (\% Top-1 accuracy) of adversarial training mechanisms (\emph{lower is better}). Our proposed attack (G-PGA, Algo. \ref{alg:guided_attack}) efficiently exposes any elusive robustness within few attack iterations.}
		\label{tab:main-results}
	\end{table*}

%% file: tables/evalutaion_with_no_label_smoothing.tex
\begin{table}[!t]
	\centering
		\setlength{\tabcolsep}{7pt}
		\scalebox{1}[1]{
		\begin{tabular} { ccccc }
			\toprule
				\rowcolor{Gray}
			Dataset & Defense & PGD & AA-Full& G-PGA  \\
			 \hline
			 \multirow{2}{*}{CIFAR10} & Madry \etal \cite{madry2018towards}  & 46.3 & \textbf{44.0}  &\textbf{44.0}\\
			 & Trades \cite{Zhang2019theoretically}  &56.4 &53.1&\textbf{52.8}\\
			\midrule
			 \multirow{2}{*}{CIFAR100} & Madry \etal \cite{madry2018towards}  & 25.3 & 24.0  &\textbf{23.0}\\
			 & Trades \cite{Zhang2019theoretically}  &27.9 &\textbf{25.1}&\textbf{25.1}\\
			 \midrule
			 \multirow{2}{*}{SVHN} & Madry \etal \cite{madry2018towards}  & 50.7 & \textbf{46.0}  &\textbf{46.0}\\
			 & Trades \cite{Zhang2019theoretically}  &60.9 & 56.8&\textbf{55.0}\\
			 \midrule
	\end{tabular}}
	\caption{ Performances comparison of different attacks including PGD, AA-Full and G-PGA against different defenses. No label smoothing is used during training of these defenses that is $\delta=0$. Top-1 (\%) accuracy is reported on the test sets of each dataset (\emph{lower is better}.)}
	\label{tab: evaluation_at_delta_0}
\end{table}

%% file: plots/ablation_whitebox_vs_blackbox_vs_ours.tex
\begin{figure*}[!t]
\centering
  \begin{minipage}{.245\textwidth}
  	\centering
    \includegraphics[ width=\linewidth,keepaspectratio, clip=true,trim=0 0 0 0mm]{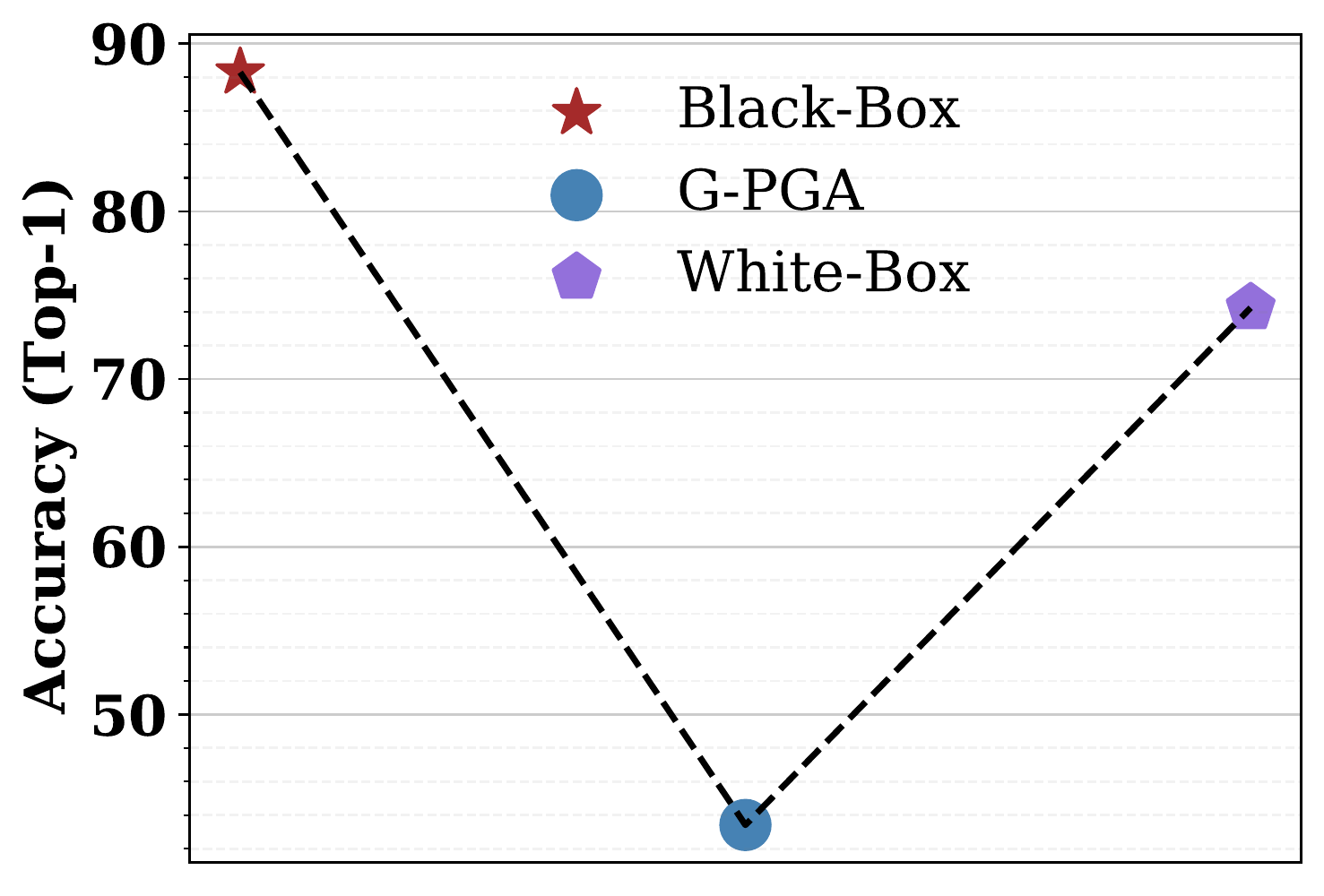}
    \tiny  (a) \textbf{ Effect of model $h$ ($\beta=0$)}
  \end{minipage}
  \begin{minipage}{.245\textwidth}
  	\centering
    \includegraphics[width=\linewidth, keepaspectratio, clip=true,trim=0 0 0 0mm]{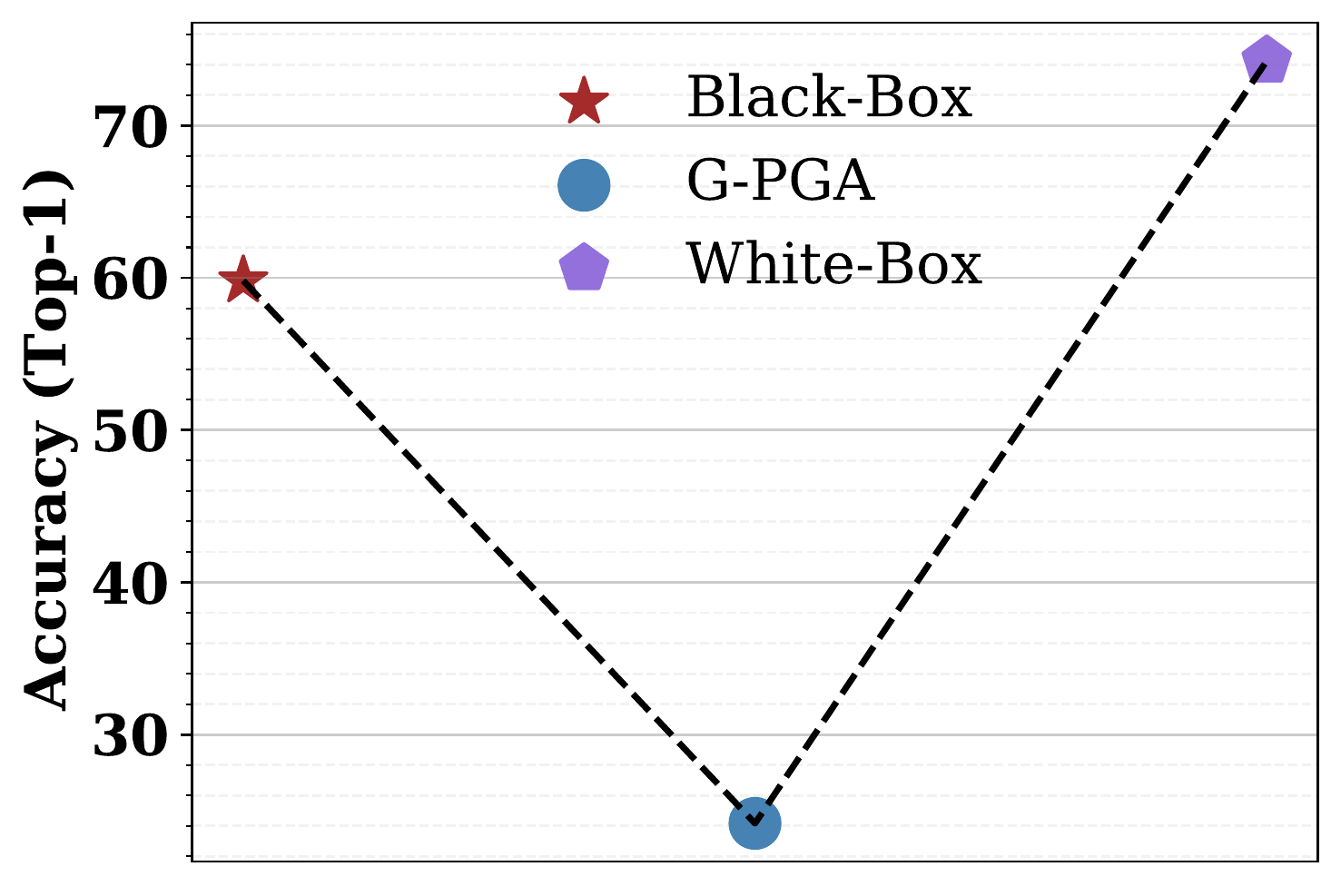}
    \tiny  (b) \textbf{Effect of model $h$ ($\beta=1$)}
  \end{minipage}
  \begin{minipage}{.245\textwidth}
  	\centering
    \includegraphics[width=\linewidth, keepaspectratio, clip=true,trim=0 0 0 0mm]{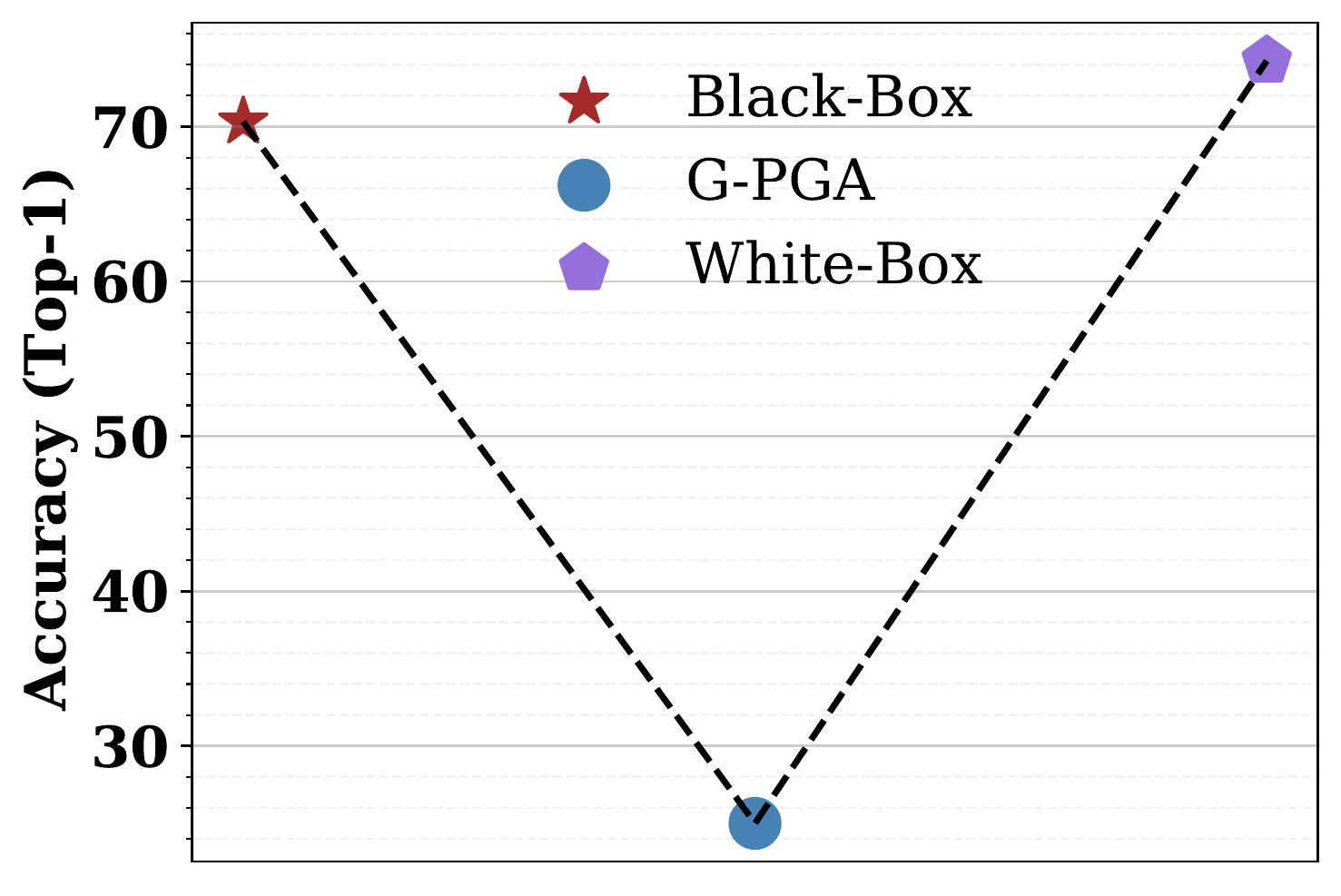}\
    \tiny  (c) \textbf{Effect of model $h$ ($\beta=6$)}
  \end{minipage}
    \begin{minipage}{.245\textwidth}
  	\centering
    \includegraphics[width=\linewidth, keepaspectratio, clip=true,trim=0 0 0 0mm]{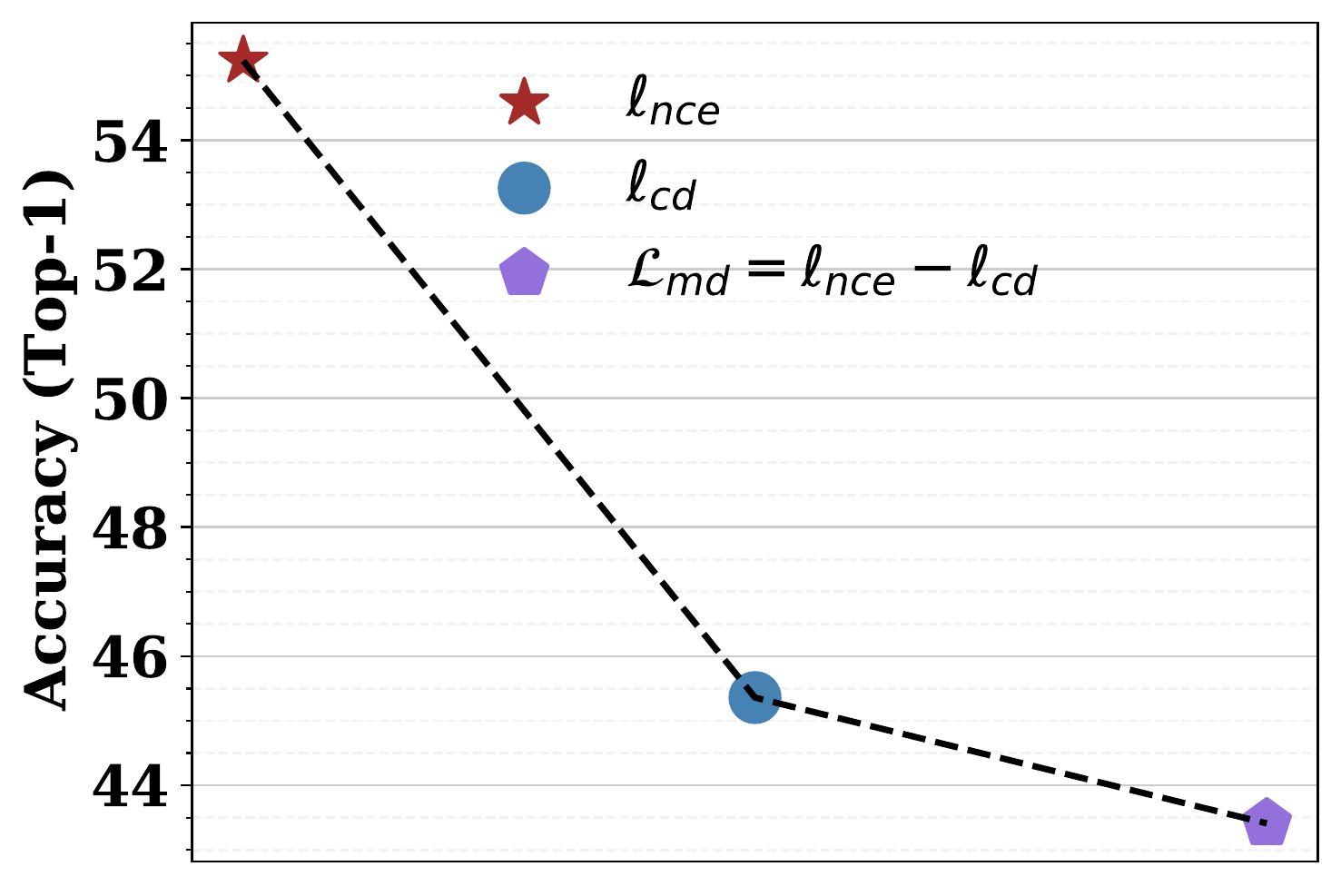}\
    \tiny  (d) \textbf{Effect of model $h$ ($\beta=0$)}
  \end{minipage}
  \caption{Accuracy (\%) of ResNet18 on CIFAR10 trained using Mask-AT (Algorithm \ref{alg:masking_at}, $\eta=6, \delta=0.75$) against different attacks. It clearly demonstrates the fooling ability  of match and deceive w.r.t black-box (weak) and white-box (inefficient due to gradient masking) attacks. Results are reported by running PGD attack with 20 iterations with step-size of $\frac{2}{255}$  (lower is better). In black-box setting, adversaries are computed on ResNet18 trained using Trades \cite{Zhang2019theoretically} with $\beta=0 \text{ (a)}, \beta=1 \text{ (b)}, \beta=6 \text{ (c)}$. Finally, plot (d) demonstrates the effectiveness of each component of match and deceive loss (Eq. \ref{eq:md}).} \vspace{-0.2cm}
\label{fig:analysis_match_and_deceive}
\end{figure*}

%% file: bare_jrnl.bbl
\begin{thebibliography}{10}
\providecommand{\url}[1]{#1}
\csname url@samestyle\endcsname
\providecommand{\newblock}{\relax}
\providecommand{\bibinfo}[2]{#2}
\providecommand{\BIBentrySTDinterwordspacing}{\spaceskip=0pt\relax}
\providecommand{\BIBentryALTinterwordstretchfactor}{4}
\providecommand{\BIBentryALTinterwordspacing}{\spaceskip=\fontdimen2\font plus
\BIBentryALTinterwordstretchfactor\fontdimen3\font minus
  \fontdimen4\font\relax}
\providecommand{\BIBforeignlanguage}[2]{{%
\expandafter\ifx\csname l@#1\endcsname\relax
\typeout{** WARNING: IEEEtran.bst: No hyphenation pattern has been}%
\typeout{** loaded for the language `#1'. Using the pattern for}%
\typeout{** the default language instead.}%
\else
\language=\csname l@#1\endcsname
\fi
#2}}
\providecommand{\BIBdecl}{\relax}
\BIBdecl

\bibitem{papernot2017practical}
N.~Papernot, P.~McDaniel, I.~Goodfellow, S.~Jha, Z.~B. Celik, and A.~Swami,
  ``Practical black-box attacks against machine learning,'' in
  \emph{Proceedings of the 2017 ACM on Asia conference on computer and
  communications security}, 2017, pp. 506--519.

\bibitem{feature_scatter}
H.~Zhang and J.~Wang, ``Defense against adversarial attacks using feature
  scattering-based adversarial training,'' in \emph{Advances in Neural
  Information Processing Systems}, 2019.

\bibitem{lee2020adversarial}
S.~Lee, H.~Lee, and S.~Yoon, ``Adversarial vertex mixup: Toward better
  adversarially robust generalization,'' \emph{arXiv preprint
  arXiv:2003.02484}, 2020.

\bibitem{Pang2020Mixup}
\BIBentryALTinterwordspacing
T.~Pang*, K.~Xu*, and J.~Zhu, ``Mixup inference: Better exploiting mixup to
  defend adversarial attacks,'' in \emph{International Conference on Learning
  Representations}, 2020. [Online]. Available:
  \url{https://openreview.net/forum?id=ByxtC2VtPB}
\BIBentrySTDinterwordspacing

\bibitem{goodfellow2014explaining}
I.~J. Goodfellow, J.~Shlens, and C.~Szegedy, ``Explaining and harnessing
  adversarial examples,'' \emph{arXiv preprint arXiv:1412.6572}, 2014.

\bibitem{tramer2017ensemble}
F.~Tram{\`e}r, A.~Kurakin, N.~Papernot, D.~Boneh, and P.~McDaniel, ``Ensemble
  adversarial training: Attacks and defenses,'' in \emph{International
  Conference on Learning Representations (ICRL)}, 2018.

\bibitem{madry2018towards}
\BIBentryALTinterwordspacing
A.~Madry, A.~Makelov, L.~Schmidt, D.~Tsipras, and A.~Vladu, ``Towards deep
  learning models resistant to adversarial attacks,'' in \emph{International
  Conference on Learning Representations}, 2018. [Online]. Available:
  \url{https://openreview.net/forum?id=rJzIBfZAb}
\BIBentrySTDinterwordspacing

\bibitem{Kurakin2016AdversarialEI}
I.~J. Goodfellow, J.~Shlens, and C.~Szegedy, ``Adversarial examples in the
  physical world,'' in \emph{International Conference on Learning
  Representations (ICRL)}, 2017.

\bibitem{xie2019improving}
C.~Xie, Z.~Zhang, Y.~Zhou, S.~Bai, J.~Wang, Z.~Ren, and A.~Yuille, ``Improving
  transferability of adversarial examples with input diversity,'' in
  \emph{Computer Vision and Pattern Recognition}.\hskip 1em plus 0.5em minus
  0.4em\relax IEEE, 2019.

\bibitem{dong2018boosting}
Y.~Dong, F.~Liao, T.~Pang, H.~Su, J.~Zhu, X.~Hu, and J.~Li, ``Boosting
  adversarial attacks with momentum,'' in \emph{Proceedings of the IEEE
  Conference on Computer Vision and Pattern Recognition}, 2018.

\bibitem{muller2019does}
R.~M{\"u}ller, S.~Kornblith, and G.~E. Hinton, ``When does label smoothing
  help?'' in \emph{Advances in Neural Information Processing Systems}, 2019,
  pp. 4696--4705.

\bibitem{pereyra2017regularizing}
G.~Pereyra, G.~Tucker, J.~Chorowski, {\L}.~Kaiser, and G.~Hinton,
  ``Regularizing neural networks by penalizing confident output
  distributions,'' \emph{arXiv preprint arXiv:1701.06548}, 2017.

\bibitem{Zhang2019theoretically}
H.~Zhang, Y.~Yu, J.~Jiao, E.~P. Xing, L.~E. Ghaoui, and M.~I. Jordan,
  ``Theoretically principled trade-off between robustness and accuracy,''
  \emph{arXiv preprint arXiv:1901.08573}, 2019.

\bibitem{tabacof2016exploring}
P.~Tabacof and E.~Valle, ``Exploring the space of adversarial images,'' in
  \emph{2016 International Joint Conference on Neural Networks (IJCNN)}.\hskip
  1em plus 0.5em minus 0.4em\relax IEEE, 2016, pp. 426--433.

\bibitem{athalye2018obfuscated}
A.~Athalye, N.~Carlini, and D.~Wagner, ``Obfuscated gradients give a false
  sense of security: Circumventing defenses to adversarial examples,''
  \emph{arXiv preprint arXiv:1802.00420}, 2018.

\bibitem{shafahi2019adversarial}
A.~Shafahi, M.~Najibi, M.~A. Ghiasi, Z.~Xu, J.~Dickerson, C.~Studer, L.~S.
  Davis, G.~Taylor, and T.~Goldstein, ``Adversarial training for free!'' in
  \emph{Advances in Neural Information Processing Systems}, 2019, pp.
  3358--3369.

\bibitem{zhang2019you}
D.~Zhang, T.~Zhang, Y.~Lu, Z.~Zhu, and B.~Dong, ``You only propagate once:
  Accelerating adversarial training via maximal principle,'' in \emph{Advances
  in Neural Information Processing Systems}, 2019, pp. 227--238.

\bibitem{zhang2019towards}
H.~Zhang, H.~Chen, C.~Xiao, S.~Gowal, R.~Stanforth, B.~Li, D.~Boning, and C.-J.
  Hsieh, ``Towards stable and efficient training of verifiably robust neural
  networks,'' \emph{arXiv preprint arXiv:1906.06316}, 2019.

\bibitem{salman2019provably}
H.~Salman, G.~Yang, J.~Li, P.~Zhang, H.~Zhang, I.~Razenshteyn, and S.~Bubeck,
  ``Provably robust deep learning via adversarially trained smoothed
  classifiers,'' \emph{arXiv preprint arXiv:1906.04584}, 2019.

\bibitem{Wong2020Fast}
\BIBentryALTinterwordspacing
E.~Wong, L.~Rice, and J.~Z. Kolter, ``Fast is better than free: Revisiting
  adversarial training,'' in \emph{International Conference on Learning
  Representations}, 2020. [Online]. Available:
  \url{https://openreview.net/forum?id=BJx040EFvH}
\BIBentrySTDinterwordspacing

\bibitem{wang2020improving}
Y.~Wang, D.~Zou, J.~Yi, J.~Bailey, X.~Ma, and Q.~Gu, ``Improving adversarial
  robustness requires revisiting misclassified examples,'' in
  \emph{International Conference on Learning Representations}, 2020.

\bibitem{carmon2019unlabeled}
Y.~Carmon, A.~Raghunathan, L.~Schmidt, J.~C. Duchi, and P.~S. Liang,
  ``Unlabeled data improves adversarial robustness,'' in \emph{Advances in
  Neural Information Processing Systems}, 2019, pp. 11\,190--11\,201.

\bibitem{guo2018countering}
\BIBentryALTinterwordspacing
C.~Guo, M.~Rana, M.~Cisse, and L.~van~der Maaten, ``Countering adversarial
  images using input transformations,'' in \emph{International Conference on
  Learning Representations}, 2018. [Online]. Available:
  \url{https://openreview.net/forum?id=SyJ7ClWCb}
\BIBentrySTDinterwordspacing

\bibitem{Akhtar_2018_CVPR}
N.~Akhtar, J.~Liu, and A.~Mian, ``Defense against universal adversarial
  perturbations,'' in \emph{The IEEE Conference on Computer Vision and Pattern
  Recognition (CVPR)}, June 2018.

\bibitem{naseer2020self}
M.~Naseer, S.~Khan, M.~Hayat, F.~S. Khan, and F.~Porikli, ``A self-supervised
  approach for adversarial robustness,'' in \emph{Proceedings of the IEEE/CVF
  Conference on Computer Vision and Pattern Recognition}, 2020, pp. 262--271.

\bibitem{zhang2017mixup}
H.~Zhang, M.~Cisse, Y.~N. Dauphin, and D.~Lopez-Paz, ``mixup: Beyond empirical
  risk minimization,'' \emph{arXiv preprint arXiv:1710.09412}, 2017.

\bibitem{papernot2016limitations}
N.~Papernot, P.~McDaniel, S.~Jha, M.~Fredrikson, Z.~B. Celik, and A.~Swami,
  ``The limitations of deep learning in adversarial settings,'' in
  \emph{EuroS\&P}, 2016.

\bibitem{modas2019sparsefool}
A.~Modas, S.-M. Moosavi-Dezfooli, and P.~Frossard, ``Sparsefool: a few pixels
  make a big difference,'' in \emph{CVPR}, 2019.

\bibitem{su2019one}
J.~Su, D.~V. Vargas, and K.~Sakurai, ``One pixel attack for fooling deep neural
  networks,'' in \emph{IEEE Transactions on Evolutionary Computation}.\hskip
  1em plus 0.5em minus 0.4em\relax IEEE, 2019.

\bibitem{carlini2017towards}
N.~Carlini and D.~Wagner, ``Towards evaluating the robustness of neural
  networks,'' in \emph{2017 ieee symposium on security and privacy (sp)}.\hskip
  1em plus 0.5em minus 0.4em\relax IEEE, 2017, pp. 39--57.

\bibitem{naseer2019cross}
M.~Naseer, S.~H. Khan, H.~Khan, F.~S. Khan, and F.~Porikli, ``Cross-domain
  transferability of adversarial perturbations,'' \emph{Advances in Neural
  Information Processing Systems}, 2019.

\bibitem{Naseer_2021_ICCV}
M.~Naseer, S.~Khan, M.~Hayat, F.~S. Khan, and F.~Porikli, ``On generating
  transferable targeted perturbations,'' in \emph{Proceedings of the IEEE/CVF
  International Conference on Computer Vision (ICCV)}, October 2021, pp.
  7708--7717.

\bibitem{naseer2019local}
M.~Naseer, S.~Khan, and F.~Porikli, ``Local gradients smoothing: Defense
  against localized adversarial attacks,'' in \emph{2019 IEEE Winter Conference
  on Applications of Computer Vision (WACV)}.\hskip 1em plus 0.5em minus
  0.4em\relax IEEE, 2019, pp. 1300--1307.

\bibitem{chen2018ead}
P.-Y. Chen, Y.~Sharma, H.~Zhang, J.~Yi, and C.-J. Hsieh, ``Ead: elastic-net
  attacks to deep neural networks via adversarial examples,'' in \emph{AAAI},
  2018.

\bibitem{wu2020skip}
D.~Wu, Y.~Wang, S.-T. Xia, J.~Bailey, and X.~Ma, ``Skip connections matter: On
  the transferability of adversarial examples generated with resnets,'' in
  \emph{ICLR}, 2020.

\bibitem{naseer2021improving}
\BIBentryALTinterwordspacing
M.~Naseer, K.~Ranasinghe, S.~Khan, F.~Khan, and F.~Porikli, ``On improving
  adversarial transferability of vision transformers,'' in \emph{International
  Conference on Learning Representations}, 2022. [Online]. Available:
  \url{https://openreview.net/forum?id=D6nH3719vZy}
\BIBentrySTDinterwordspacing

\bibitem{tashiro2020output}
Y.~Tashiro, Y.~Song, and S.~Ermon, ``Output diversified initialization for
  adversarial attacks,'' \emph{arXiv preprint arXiv:2003.06878}, 2020.

\bibitem{croce2020reliable}
F.~Croce and M.~Hein, ``Reliable evaluation of adversarial robustness with an
  ensemble of diverse parameter-free attacks,'' \emph{arXiv preprint
  arXiv:2003.01690}, 2020.

\bibitem{andriushchenko2019square}
M.~Andriushchenko, F.~Croce, N.~Flammarion, and M.~Hein, ``Square attack: a
  query-efficient black-box adversarial attack via random search,'' \emph{arXiv
  preprint arXiv:1912.00049}, 2019.

\bibitem{croce2020minimally}
\BIBentryALTinterwordspacing
F.~Croce and M.~Hein, ``Minimally distorted adversarial examples with a fast
  adaptive boundary attack,'' 2020. [Online]. Available:
  \url{https://openreview.net/forum?id=HJlzxgBtwH}
\BIBentrySTDinterwordspacing

\bibitem{tramer2020adaptive}
F.~Tramer, N.~Carlini, W.~Brendel, and A.~Madry, ``On adaptive attacks to
  adversarial example defenses,'' \emph{arXiv preprint arXiv:2002.08347}, 2020.

\bibitem{pang2021bag}
\BIBentryALTinterwordspacing
T.~Pang, X.~Yang, Y.~Dong, H.~Su, and J.~Zhu, ``Bag of tricks for adversarial
  training,'' in \emph{International Conference on Learning Representations},
  2021. [Online]. Available: \url{https://openreview.net/forum?id=Xb8xvrtB8Ce}
\BIBentrySTDinterwordspacing

\bibitem{shafahi2019label}
A.~Shafahi, A.~Ghiasi, F.~Huang, and T.~Goldstein, ``Label smoothing and logit
  squeezing: a replacement for adversarial training?'' \emph{arXiv preprint
  arXiv:1910.11585}, 2019.

\bibitem{fu2020label}
C.~Fu, H.~Chen, N.~Ruan, and W.~Jia, ``Label smoothing and adversarial
  robustness,'' \emph{arXiv preprint arXiv:2009.08233}, 2020.

\bibitem{lee2020gradient}
H.~Lee, H.~Bae, and S.~Yoon, ``Gradient masking of label smoothing in
  adversarial robustness,'' \emph{IEEE Access}, vol.~9, pp. 6453--6464, 2020.

\bibitem{paszke2019pytorch}
A.~Paszke, S.~Gross, F.~Massa, A.~Lerer, J.~Bradbury, G.~Chanan, T.~Killeen,
  Z.~Lin, N.~Gimelshein, L.~Antiga \emph{et~al.}, ``Pytorch: An imperative
  style, high-performance deep learning library,'' \emph{Advances in neural
  information processing systems}, vol.~32, 2019.

\bibitem{salman2020adversarially}
H.~Salman, A.~Ilyas, L.~Engstrom, A.~Kapoor, and A.~Madry, ``Do adversarially
  robust imagenet models transfer better?'' in \emph{ArXiv preprint
  arXiv:2007.08489}, 2020.

\bibitem{Pang_github}
T.~Pang, ``Mixup-inference,'' \url{https://github.com/P2333/Mixup-Inference},
  2019.

\bibitem{deng2009imagenet}
J.~Deng, W.~Dong, R.~Socher, L.-J. Li, K.~Li, and L.~Fei-Fei, ``Imagenet: A
  large-scale hierarchical image database,'' in \emph{2009 IEEE conference on
  computer vision and pattern recognition}.\hskip 1em plus 0.5em minus
  0.4em\relax Ieee, 2009, pp. 248--255.

\bibitem{byun2022effectiveness}
J.~Byun, H.~Go, and C.~Kim, ``On the effectiveness of small input noise for
  defending against query-based black-box attacks,'' in \emph{Proceedings of
  the IEEE/CVF Winter Conference on Applications of Computer Vision}, 2022, pp.
  3051--3060.

\bibitem{qin2021random}
\BIBentryALTinterwordspacing
Z.~Qin, Y.~Fan, H.~Zha, and B.~Wu, ``Random noise defense against query-based
  black-box attacks,'' in \emph{Advances in Neural Information Processing
  Systems}, A.~Beygelzimer, Y.~Dauphin, P.~Liang, and J.~W. Vaughan, Eds.,
  2021. [Online]. Available: \url{https://openreview.net/forum?id=ZPSD4xZc6j8}
\BIBentrySTDinterwordspacing

\bibitem{ilyas2019adversarial}
A.~Ilyas, S.~Santurkar, D.~Tsipras, L.~Engstrom, B.~Tran, and A.~Madry,
  ``Adversarial examples are not bugs, they are features,'' \emph{Advances in
  neural information processing systems}, vol.~32, 2019.

\bibitem{selvaraju2017grad}
R.~R. Selvaraju, M.~Cogswell, A.~Das, R.~Vedantam, D.~Parikh, and D.~Batra,
  ``Grad-cam: Visual explanations from deep networks via gradient-based
  localization,'' in \emph{Proceedings of the IEEE international conference on
  computer vision}, 2017, pp. 618--626.

\end{thebibliography}
